\pdfoutput=1
\documentclass[11pt]{article}

\usepackage[]{acl}

\usepackage{times}
\usepackage{latexsym}

\usepackage[T1]{fontenc}
\usepackage[utf8]{inputenc}

\usepackage{microtype}

\usepackage{multirow}
\usepackage{subcaption}
\usepackage{amssymb}
\usepackage{amsmath}
\usepackage{bbm}
\usepackage{enumitem}
\usepackage{float} % for the tables...
\usepackage[flushleft]{threeparttable}
\usepackage{arydshln}

\usepackage{scalerel,graphicx,xparse}
\NewDocumentCommand\emojiherb{}{\scalerel*{\includegraphics{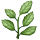}}{X}}

\title{HERB$^{\emojiherb{}{}}$: Measuring Hierarchical Regional Bias in \\ Pre-trained Language Models}

\newcommand*\samethanks[1][\value{footnote}]{\footnotemark[#1]}

\author{ 
    \small
    Yizhi Li\textsuperscript{1}\thanks{\quad The two authors contributed equally to this work.}\space, 
    Ge Zhang\textsuperscript{2\space 3\space 4}\samethanks\space, 
    Bohao Yang\textsuperscript{1},
    Chenghua Lin\textsuperscript{1}\thanks{\quad Corresponding authors.}\;,
    Shi Wang\textsuperscript{3}\samethanks\space, 
    Anton Ragni\textsuperscript{1},   
    Jie Fu\textsuperscript{2} \\ 
{\small 
\textsuperscript{1} Department of Computer Science, The University of Sheffield, UK
}\vspace{-1.5mm} \\ 
{\small 
\textsuperscript{2} Beijing Academy of Artificial Intelligence, China
}\vspace{-1.5mm} \\
{\small 
\textsuperscript{3} Institute of Computing Technology, Chinese Academy of Sciences, China
}\vspace{-1.5mm} \\
{\small 
\textsuperscript{4} University of Michigan Ann Arbor, USA
}\vspace{-1.5mm} \\
\texttt{\small
\{yizhi.li, byang27, c.lin, a.ragni\}@sheffield.ac.uk\textsuperscript{1}, 
}\vspace{-1.5mm} \\
\texttt{\small
gezhang@umich.edu\textsuperscript{2}, wangshi@ict.ac.cn\textsuperscript{3}, fujie@baai.ac.cn\textsuperscript{4}
}
}
\begin{document}
\maketitle

\begin{abstract}
\textcolor{teal}{Content Warning: This work contains examples that potentially implicate stereotypes, associations, and other harms that could be offensive to individuals in certain regions.}

\noindent Fairness has become a trending topic in natural language processing (NLP), which addresses biases targeting certain social groups such as genders and religions. 
However, regional bias in language models (LMs), a long-standing global discrimination problem, still remains unexplored.
%Consequently, we
This paper bridges the gap by analysing the regional bias learned by the pre-trained language models that are broadly used in NLP tasks. 
In addition to verifying the existence of regional bias in LMs, we find that the biases on regional groups can be strongly influenced
%largely affected 
by the geographical clustering of the groups.
We accordingly propose a HiErarchical Regional Bias evaluation method (HERB$^{\emojiherb{}{}}$) utilising the information from the sub-region clusters to quantify the bias in pre-trained LMs.
Experiments show that our hierarchical metric can effectively evaluate the regional bias with respect to comprehensive topics and measure the potential regional bias that can be propagated to downstream tasks.
Our codes are available at \url{https://github.com/Bernard-Yang/HERB}.
%Our codes are available at \href{https://github.com/Bernard-Yang/HERB}{\color{blue}{https://github.com/Bernard-Yang/HERB}}.
% Last but not least, we not only conduct extensive bias evaluations on the state-of-the-art pre-trained language models but also provide a technique for regional bias mitigation.
\end{abstract}

\section{Introduction}
% Para1
Large-scale pre-trained language models (LMs) are prevalent in the natural language processing (NLP) community since the costly pre-trained models can be adapted to a wide range of downstream applications. 
%social group definition
% A \textit{social group} refers to a group of the population who share a common social identity and similar mutual expectations according to the demographic studies~\cite{tajfel2010social}.
However, research studies demonstrate that the societal biases in the pre-training corpora can be learned by LMs and further propagated to the downstream applications~\cite{zhao-etal-2019-gender, dev2020measuring, goldfarb2020intrinsic, kurita_measuring_2019}. 
% (people grouped by demographic characteristics~\cite{tajfel2010social})
To qualify and mitigate bias for pre-trained LMs, researchers have developed bias evaluation methods targeting certain \textit{social groups} such as gender, religion, and race~\cite{sun-etal-2019-mitigating, manzini_black_2019, xia-etal-2020-demoting, delobelle_measuring_2021}.
% although the societal biases on social groups categorised by regions appear frequently, 
However, existing methods do not examine the social groups categorised by geographical information, which leaves the region-related biases in pre-trained LMs unexplored. 
Therefore, our work bridges this gap by addressing research questions about whether regional bias exists in the pre-trained LMs, and if yes, how to quantify the bias in a principled way.
%and the  of such a bias.

% Except for stud, the regional bias and the hierarchy of social group categorisation.
% \citet{walker2015personal} shows that social bias can be introduced in the single-blind academic peer review system when the geographical characteristic of the authors are given.

\begin{figure}[!tb]
\centering
{
\centering
\begin{subfigure}{0.44\linewidth}
\centering
\includegraphics[width=\linewidth]
{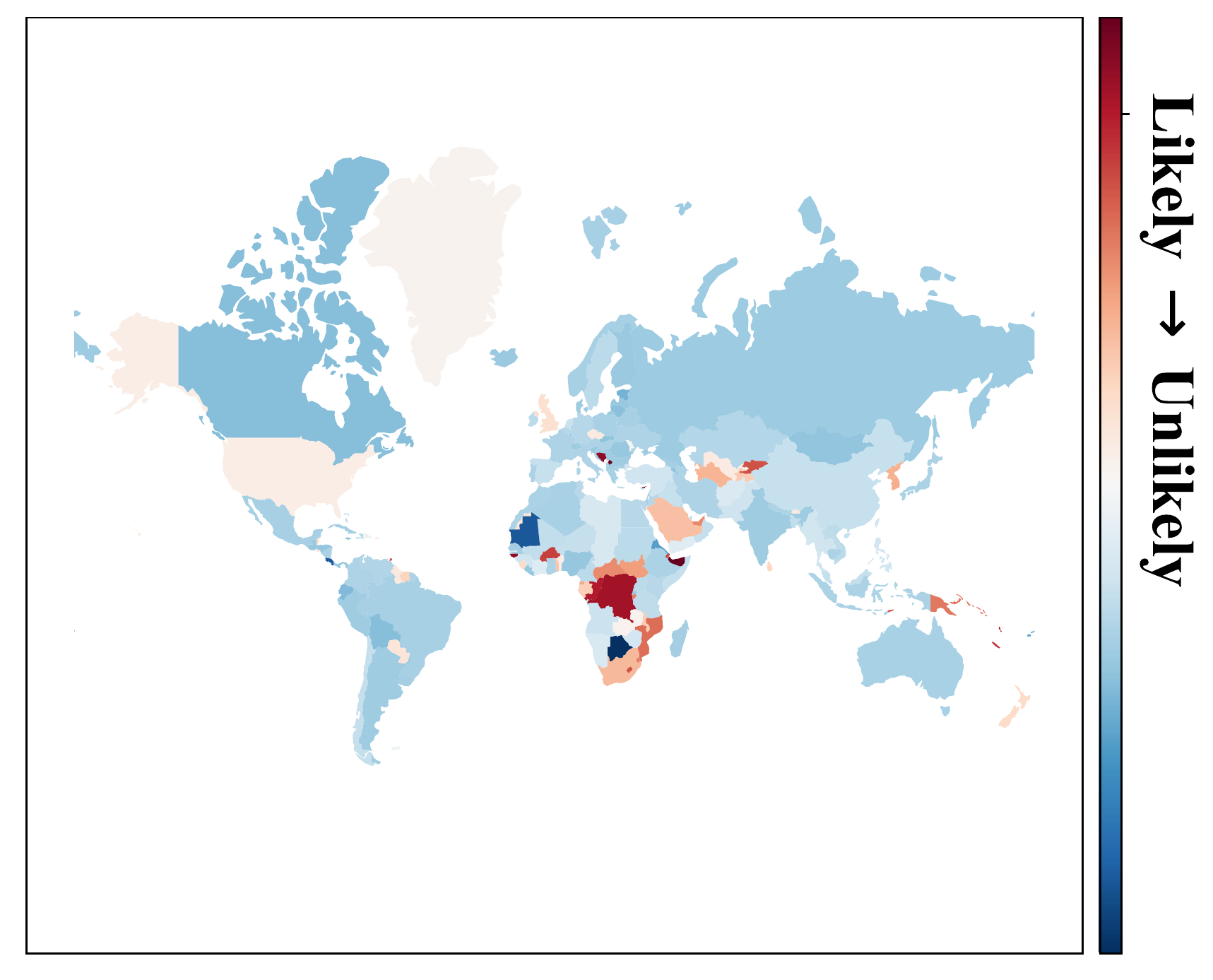} 
\caption{Regions Worldwide}\label{fig:descriptive_v_a}
\end{subfigure}
\begin{subfigure}{0.44\linewidth}
\centering
\includegraphics[width=\linewidth]{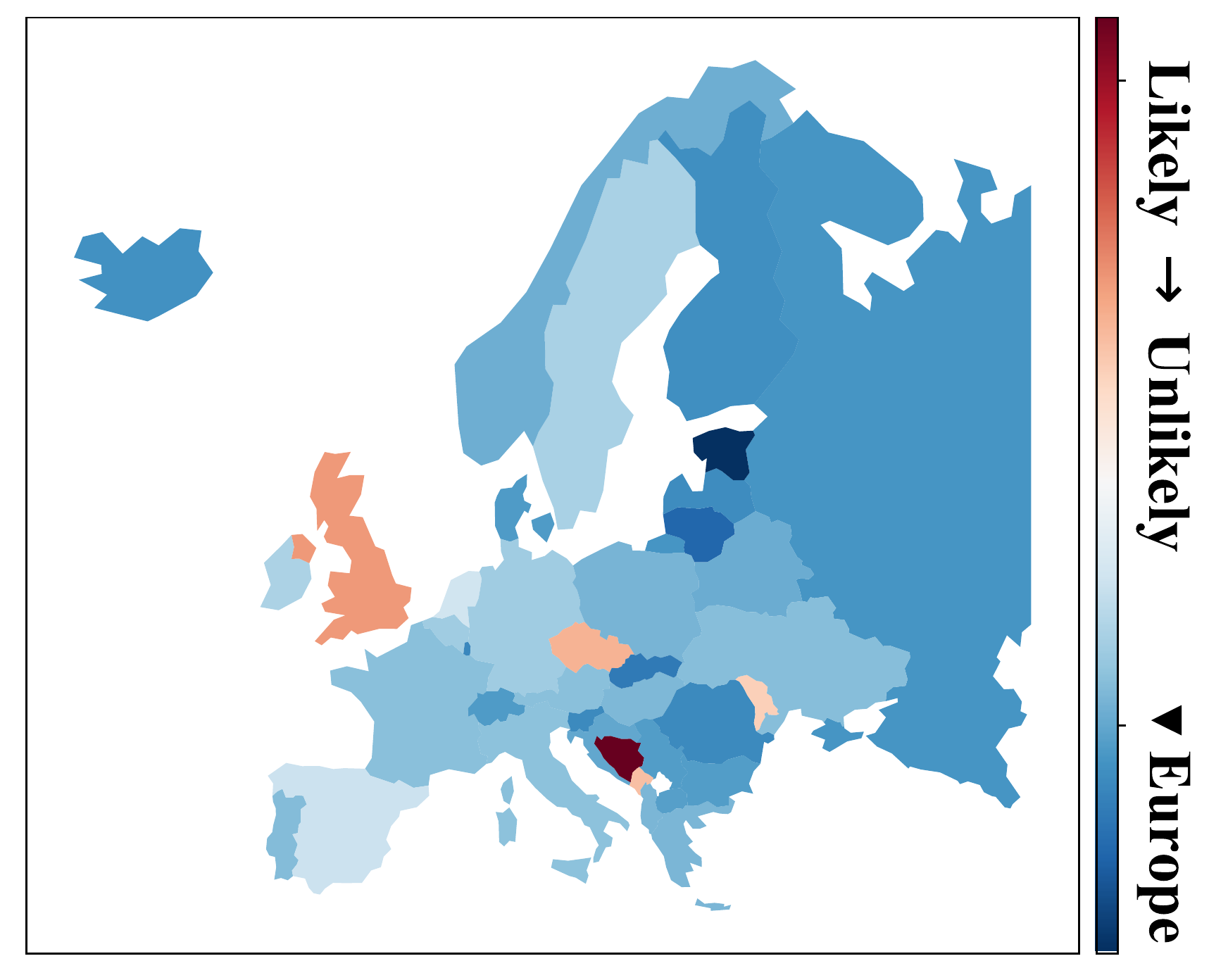} 
\caption{Regions in Europe}\label{fig:descriptive_v_b}
\end{subfigure}
}
\caption{
The Regional Likelihood in \texttt{\small [bald]} Dimension Produced by RoBERTa.
The regional likelihoods are produced with sentences filled with different region names and the fixed descriptive word \texttt{\small [bald]} in the given template.
The likelihood calculated with the region word \texttt{\small [Europe]} is marked by $\blacktriangleleft$ at the likelihood legend in Fig.~\ref{fig:descriptive_v_b}.
The details of calculation can be referred to \S\ref{sec:method_eval_descriptive_v}.
}
\label{fig:descriptive_v}
\end{figure}

% Para2: regional bias is existing 
Bias in NLP applications makes distinct judgements on people based on their gender, race, religion, region, or other social groups could be harmful, such as automatically downgrading the resumes of female applicants in recruiting~\cite{Dastin2018amazonRecruiting}
% \textcolor{red}{while failing to study people only regards the context such as automatically downgrading the resumes from female applicants in recruiting}~\cite{Dastin2018amazonRecruiting}.
Regional bias represents stereotypes based on the geographic location where people live or come from \cite{wiki:Discrimination}. To verify the existence of regional bias, we first leverage a sentence-level bias measurement~\cite{kaneko2022unmasking}, with which the likelihood of a biased sentence produced by a pre-trained LM can be acquired with a designed input:

\vspace{1mm}
\centerline{
\small {People in \texttt{[region]} are \texttt{[description]}.}
}
\vspace{1mm}

\noindent where \texttt{\small [region]} and \texttt{\small [description]} can be filled with any desired words. 
The output likelihood represents the contextualised possibility of associating people in the region with the given context, which can be utilised to analyse the bias integrated into LMs.
%, which could be caused by the distortion of the training corpora.
% \textcolor{blue}{limitation: opinion perspectives, the distortion is caused by speakers, population ratio...}
From the perspective of the pre-trained LM, there is a `world map' of region-wide judgements regards to the \texttt{\small [description]} of interest. 
As the case shown in Fig.~\ref{fig:descriptive_v}, the pre-trained RoBERTa~\cite{liu2019roberta} holds a prejudice that people in specific regions are more likely to be \texttt{\small [bald]}, which hardly stands for the facts and could amplify the regional bias.

% Para3: regional bias is hierarchical
In addition, we discover that the regional bias in pre-trained LMs could be hierarchical as demonstrated in Fig.~\ref{fig:descriptive_v_b}.
Whilst people in many European countries share a low likelihood of \texttt{\small [bald]}, the upper-level regional group, i.e., Europe, is also assigned a relatively small likelihood.
This suggests that the language models do recognise the hierarchical structure of the regional group structure and thus produce similar results for most of the countries and the continental group.
However, opposite trends of high likelihoods appear in countries such as the United Kingdom, which implies that bias in these regions could not be represented by the higher-level group, Europe. 
Without considering relationships between regional groups, the modelling of regional bias is difficult because only conducting bias evaluation on high-level groups can disguise the biases in their sub-regions.
% It makes the evaluation of regional bias even more challenging that most of the existing bias evaluation methods only consider the relationship between the target social groups and specific kinds of bias rather than the group-group relationships.

% complicated bias-bias 要提吗

To tackle the aforementioned issues, we argue that the design of regional bias evaluation for pre-trained LMs should satisfy the following criteria:
\begin{enumerate}[noitemsep] % itemsep=0.5mm
    \item The metric should leverage the structural information from sub-regions to evaluate the bias for higher-level regions.
    \item The disperancy of judgements on different regional groups in the same level should also be considered bias, e.g., inconsistent judgements on the cities in the same country.
    % E.g., inconsistent judgements are assigned to different European countries.
    % \item The descriptive words used to evaluate regional bias should cover a variety of topics.
\end{enumerate}

With the criteria in mind, we  design a clustering-based metric \textbf{HERB}$^{\emojiherb{}{}}$, which can effectively measure \textbf{H}i\textbf{E}rarchical \textbf{R}egional \textbf{B}ias. %for the regional bias 
% with respect to an abundant descriptive word list.
HERB$^{\emojiherb{}{}}$ is grounded on the \textit{descriptive vectors}, a novel component that is designed to capture region-specific
%\textcolor{orange}{We first calculate the} \textit{descriptive vectors} for each region as the basis of our method, which contains
contextualised likelihoods with respect to the content of  \texttt{\small [description]}.
% The abundant descriptive word list includes adjectives and nouns describing occupation, intelligence, appearance, strength, and morality.
As the bias on regions should be relevant to their sub-region, we formalise the bias on a given region as the \textit{sparseness} of its sub-region cluster in the descriptive space.
% which uses the pairwise distance of the sub-region descriptive vectors to represent the bias of the upper-level regions.
The intuition behind the cluster-based sparseness calculation is that the more bias exists in the region, the more inconsistent the judgements on its sub-regions received. % from the pre-trained language models
In the case that a region does not contain any sub-regions, its cluster sparseness is modelled by the distance to the centroids of the cluster, where all the regions belong to the same upper-level region, e.g., cities in the same country.
% can be represented by how inconsistent the judgements from the pre-trained language models on its sub-regions are. 
% Moreover, while utilising the semantic hierarchical information of the regional groups, 
We further propose aggregation functions for the descriptive vector and cluster-based bias calculation to utilise the hierarchy. The aggregated cluster-based bias evaluation not only empowers our metric to consider regional bias at multiple levels but also sheds light on the general regional bias evaluation for the pre-trained LMs.
% contained in the given pre-trained language model.
% not only simultaneously detect the biases in relevant social groups but also to \textcolor{blue}{[TODO] do something else special}.
% we further propose aggregation functions

% In our experiments, w
We perform extensive evaluations of hierarchical regional bias on various state-of-the-art pre-trained language models 
% with various sizes of parameters. 
and study the regional hierarchical relationships learned by the LMs.
% and prove the effectiveness of our aggregated functions.
% \textcolor{blue}{explain the bald countries}
% downstream task
Additionally, we conduct experiments to study the propagation of regional bias from pre-trained models to downstream tasks. 
By introducing extra neutral regional information to the test samples and observing the prediction change, we evaluate how much the model performances are affected by region bias.
% using the numbers of changed predictions. 
Regional bias evaluation results on downstream tasks confirm that results from our metric have correlations to the bias propagation to fine-tuned LMs. 

% \textcolor{blue}{implicate the harm to downstream tasks.}
% In conclusion, we propose an evaluation method to quantify the regional bias in pre-trained LMs, while considering the hierarchical regional group structure and measuring the bias with selective description words.
% from a wide range of aspects defined by selective description words.

% \clearpage
\section{Related Work}

\textbf{Regional bias} has been recognised as one of the main concerns of the United Nations \cite{ramcharan2019equality}. 
Its severe influence has been detected and verified in various areas, including scientific research \cite{paris1998region}, economics \cite{ramcharan2019equality}, agriculture \cite{jia2022much}, customer satisfaction investigation \cite{ibeke-etal-2017-extracting,brint2021regional}, and public opinion \cite{peng2021amplification}.
Extensive regional bias is often decomposed into national and regional biases \cite{paris1998region, jia2022much, SAARINEN20211214}, which inspires us to consider designing the metric of regional biases in the language models(LMs) hierarchically. 

\noindent\textbf{Societal biases in NLP} has raised increasing attention because large-scale LMs containing societal biases can produce undesirable biased expressions and have negative societal impacts on the minorities \cite{sheng2021societal}.  
Existing natural language processing researchers have detected and analysed regional bias against people in specific areas \cite{abid2021persistent,sheng2021societal}. 
But there is still no well-formalized metric for regional bias contained in LMs, like gender bias \cite{bordia-bowman-2019-identifying, sheng-etal-2019-woman}, racial bias \cite{solaiman2019release, groenwold-etal-2020-investigating}, political bias \cite{liu2021mitigating}, religious bias \cite{abid2021persistent}, and profession bias \cite{huang-etal-2020-reducing}. 

\noindent\textbf{Societal bias metrics} include regard ratio \cite{sheng-etal-2019-woman}, sentiment ratio \cite{groenwold-etal-2020-investigating}, individual and group fairness \cite{huang-etal-2020-reducing}, and word co-occurrence score \cite{bordia-bowman-2019-identifying}.   
Additionally, societal bias is also classified based on how human detects it in the corpus. 
\citeauthor{liu2021mitigating} classifies societal bias into direct bias and indirect bias, based on whether measures bias of texts generated using prompts with ideological triggers. 
Societal bias in texts can also be classified into contextual-level societal bias \cite{bartl2020unmasking} and word-level societal bias \cite{bordia-bowman-2019-identifying}, based on how it is detected from texts. 
Additionally, various well-designed word lists and perspective descriptions are used to measure societal bias. 
\citeauthor{chaloner-maldonado-2019-measuring} propose 5 target word categories, including career vs family, maths vs arts, science vs arts, intelligence vs arts, and strength with weakness, to measure gender bias in word embeddings.
\citeauthor{liu2021mitigating} propose several political topics related prompts to measure societal bias.
\citeauthor{jiao-luo-2021-gender} propose an adjective list to measure descriptive gender bias hidden in Chinese LMs.
\citeauthor{zhou2019examining} use gender-related grammar words and occupation-related words to measure gender bias.
In sharp contrast, HERB$^{\emojiherb{}{}}$ focuses on measuring contextual-level regional indirect bias. 

\section{Methodology}\label{sec:method_eval}

% \textcolor{blue}{TODO: Summarising this section.}
We describe our hierarchical evaluation method for regional bias in pre-trained LMs in this section.
To measure the bias from comprehensive aspects, we first map all the regional groups to a descriptive representation space with a selective word list.
We use a cluster-based evaluation method to represent the bias of a given region with regard to its sub-regions, which leverages the natural hierarchical regional group structure in the bias evaluation.
In order to summarise bias information from regions at different levels simultaneously, we design a novel aggregation function of the descriptive vector and cluster-based bias, which measures the general regional bias in the pre-trained LMs.
% To evaluate the regional bias with
% \subsection{Automatic Hierarchical Topic Targeting}
% We target the topics with most biases following the method proposed in~\citet{grootendorst2022bertopic}.
% We assume that we have a word list that describes the most common stereotypes and biases, like different adjectives and some actions. Then we can project the word embedding of German, English, or such into those words.
% After that, we can calculate the intrinsic bias or tear apart based on calculating cluster density or such.
% Furthermore, after a formal definition of intrinsic bias or degree of disruption, we can hierarchically combine those intrinsic values into a final value.
\subsection{Descriptive Vector of Regions}\label{sec:method_eval_descriptive_v}

To quantify the judgements on a given regional social group, we design a descriptive vector $v$ which can be utilised to measure the bias from language models for each region $r$.

We collect a descriptive word list ($D = \{d_1,d_2,...,d_n\}$) containing adjectives and occupations that could show stereotypes or biases when describing people. 
% The descriptive adjectives are selected according to specific \textit{supersenses} \textcolor{blue}{[proposed by]}~\cite{tsvetkov2014augmenting}, i.e., those adjectives depicting intelligence, appearance, strength, and morality.
The adjective list depicting intelligence, appearance, and strength is from the work of~\citet{chaloner-maldonado-2019-measuring}.
To augment the list, we also apply the adjective list depicting morality from \cite{shahid-etal-2020-detecting}. We slightly modify the adjectives so that they match the prompt, and change the original list to make the size balanced across different topics.
Additionally, we include the occupation word list from \cite{Bolukbasi-2016-Man-Programmer} as part of the word list.
Because the occupation word list is adapted to a comparable size to other lists, we can use the full word list to model bias balanced on different topics.
The complete description word list is given in Appendix~\ref{sec:appendix_wordlist}. 

%, where $d_i$ represents the word embedding of the descriptive adjective word. 
% \textcolor{red}{[To Decide] whether should we do a dimension reduction process here.}

In order to conduct an in-depth analysis of the regional bias of language models, we select the regional entities at the continent, country\footnote{
The `country' does not refer to the actual sovereign states but the region concepts that are categorised as one level higher than the cities in the package \href{https://github.com/yaph/geonamescache}{geonamecache}.}, and city levels. The region word list is noted as $R = \{r_1,r_2,...,r_m\}$.
%, where $r_j$ represents the word embeddings of the regions. 
To learn the regional bias at the contextualised level, we design a template input $S_{ij}$ for language models to calculate the regional bias score for a specific region-description pair  $(d_i,r_j)$:
% Given a region-description pair $(d_i,r_j)$, we

\vspace{2mm}
\centerline{
\small {People in \texttt{[region]} are \texttt{[description]}.}
}
\vspace{2mm}

\noindent where \texttt{\small [region]} and \texttt{\small [description]} refers to the region word $r_j$ in $R$ and descriptive word $d_i$ in $D$, respectively.

Inspired by the recently proposed unmasking sequence likelihood~\cite{kaneko2022unmasking}, we use the template input $S_{ij}$ to calculate the contextualised likelihood for the given region-description pair $(d_i,r_j)$:
\begin{equation}\label{eq:f}
\small
    f(S_{ij}) = \frac{1}{|S_{ij}|}\sum_{t=1}^{|S_{ij}|}\log(P(w_t|S_{ij};\theta))
\end{equation}
where $\theta$ refers to parameters of a specific language model. 
The $f(S_{ij})$ uses the contextualised likelihood to represent how possible the pre-trained language model would think people in \texttt{\small [region]} are in connection with the word 
\texttt{\small [description]}.
% \textcolor{orange}{what if the LM is not MLM? Eg., can we use the template for the conditional input of seq2seq LMs?}

Given a region $r_j$, we can summarise the regional bias from a language model by defining the corresponding $L_2$ normalised \textit{descriptive vector}: 
\begin{align}\label{eq:v}
\small
& v'(r_j) = (f(S_{1j}),...,f(S_{nj})) \notag \\
& v(r_j) = \frac{v'(r_j)}{ ||v'(r_j)||}
\end{align}
As each of the 112 dimensions of the descriptive vector represents the judgement in a specific aspect on $r_j$, we can utilise $v(r_j)$ to measure the learned bias in language models for the given regional social group. The full list of selected descriptive words is given in Appendix~\ref{sec:appendix_wordlist}.
% \textcolor{green}{[Vector-based representation allows you to do more than just measuring.]}

\subsection{Cluster-based Regional Bias}\label{sec:method_eval_compact_c}
\begin{figure}[!bt]
\centering

{\centering
\begin{subfigure}{0.45\linewidth}
\centering
\includegraphics[width=\linewidth]{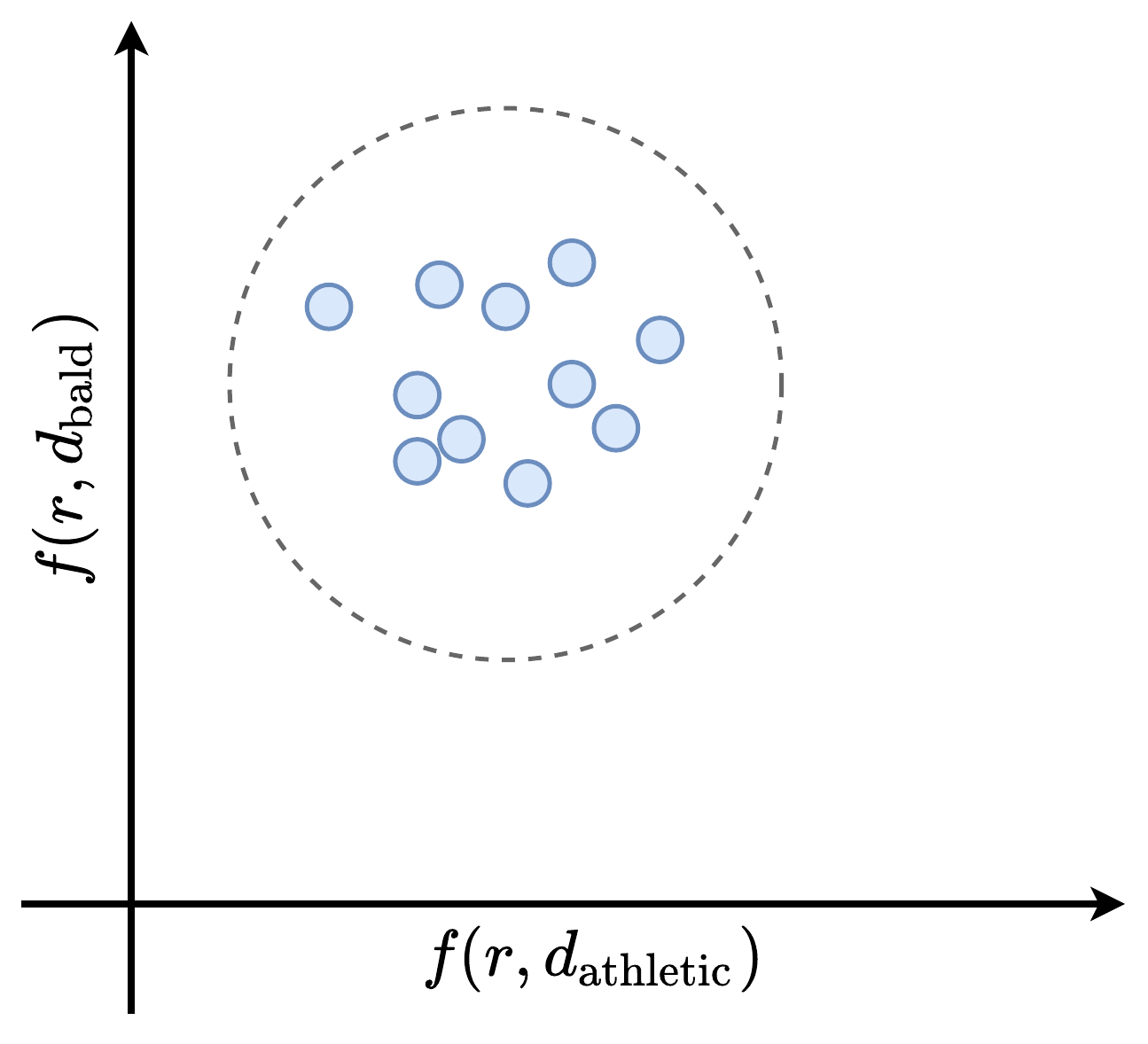} 
\caption{Low Sparseness}\label{fig:compact_high}
\end{subfigure}
\begin{subfigure}{0.45\linewidth}
\centering
\includegraphics[width=\linewidth]{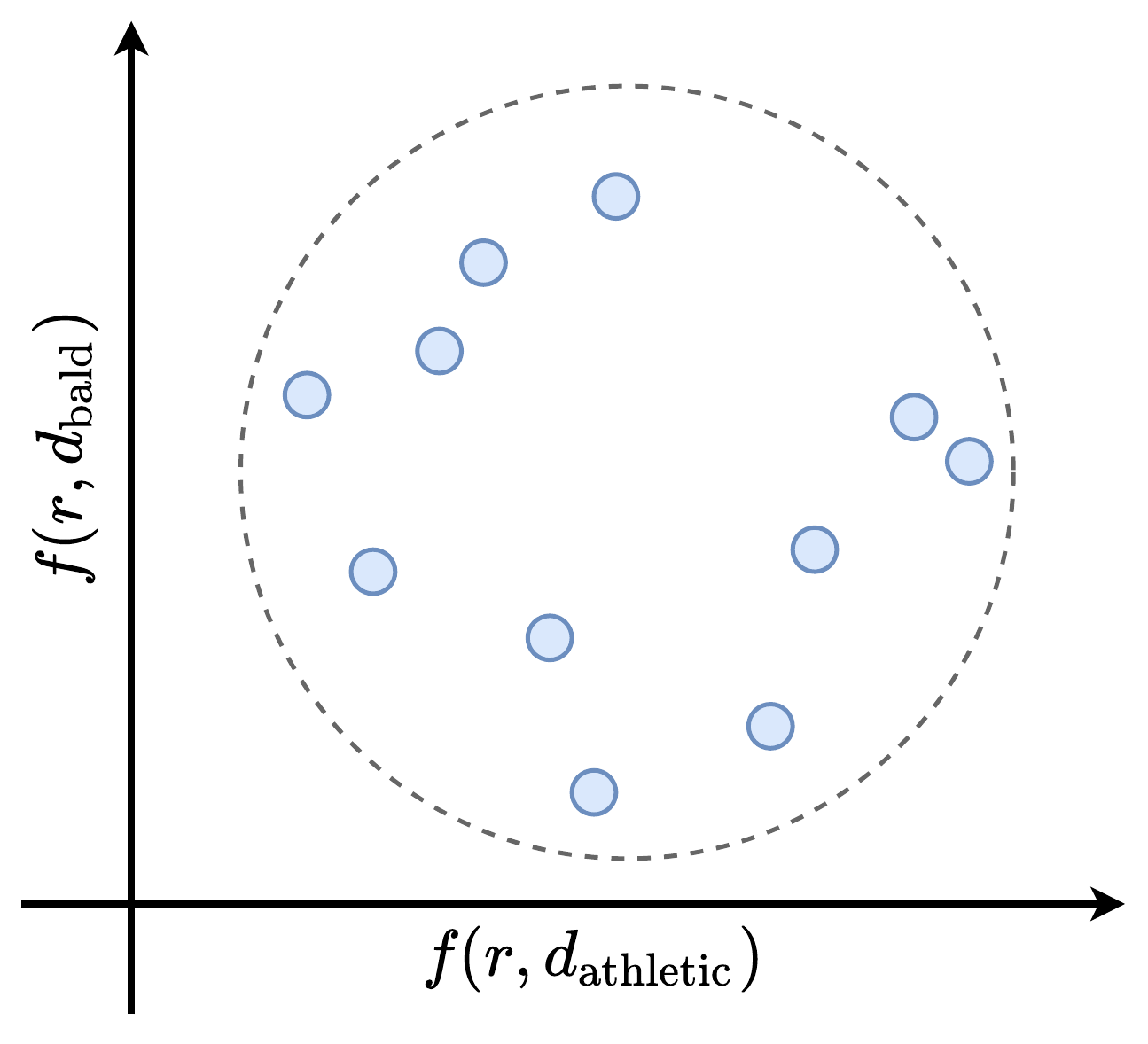} 
\caption{High Sparseness}\label{fig:compact_low}
\end{subfigure}
}
\caption{
Cluster-based Sparseness of Regional Descriptive Vectors. We show an example case when the descriptive vectors (blue dots) are two-dimensional, i.e., only calculated through two description words.
}
\label{fig:compact}
\end{figure}

Based on the natural or executive partition of the regions, we can further define clusters of regional social groups in the descriptive space with respect to a specific language model.
For example, the continent of Europe can be represented as a cluster of descriptive vectors of European countries including Germany, France, and so on.
Following the literature, we use the notation $r_k \trianglelefteq r_j$ to represents that a sub-regions $r_k$ at the lower level $l-1$ is contained inside the region $r_j$ at the higher level $l$. We thus can formalise the set of all the sub-regions included in $r_j$ as the notation $R_{\trianglelefteq r_j}$ and the set of all sub-regions $r_k$ in the same upper region as $R_{r_k \trianglelefteq}$.

We propose to use the \textit{sparseness} of a sub-region cluster to represent the inconsistency of judgements from language models. 
Intuitively, if the descriptive vectors of sub-regions are distributed further from each other, the language model would be considered to have more bias on their parent regions since the social groups inside a \textit{sparse} cluster receive distinct judgements. 
For instance, compared to Fig.~\ref{fig:compact_low}, the descriptive vectors of the cluster in Fig.~\ref{fig:compact_high} are generally closer to each other and thus the cluster is regarded as a more \textit{compact} one, which suggests the language model used to acquire the cluster contains less regional bias.

% \textcolor{green}{$R'$ looks ugly, why do you need prime here?}
The formal calculation of the sparseness $c$ of any cluster $R$ of sub-regions is defined by the average pairwise euclidean distance between the descriptive vectors:
\begin{equation}\label{eq:compact}
\small
    c(R) =  \frac{2}{|R|(|R|-1)} \cdot \sum_{r_{j1},r_{j2} \in R} ||v(r_{j1})-v(r_{j2})||
\end{equation}
It can be observed in Eq.~\ref{eq:compact} that the pairwise $L_2$ distances of descriptive vectors $v(r_{j1})$ and $v(r_{j2})$ have a direct effect on the sparseness of the given region cluster, which could be further utilised in the evaluation of the general regional bias of a language model. 
% \textcolor{green}{[Is this a classic measure of the spread of a population? Any issues to flag up?]}

\subsection{Hierarchical Regional Bias}\label{sec:method_eval_aggregation}
Since the concepts of regions are naturally partitioned and grouped by their geographic or executive administration, we state that the modelling of a region can be significantly affected by the sub-regions it contains. 
% \textcolor{blue}{TODO: maybe design an experiment to show the hierarchy}.
As a result, we define aggregation functions to leverage the hierarchical information to describe and evaluate the bias on regions in higher levels, which summarises the descriptive information and cluster-based bias from sub-regions in the lower level.

We first provide the aggregation function of the descriptive vector defined in \S\ref{sec:method_eval_descriptive_v} for a given region group $r_j$ in layer $l$: 
% \begin{align}
    % v(r_j) = v(r_j)^{(l)} + \frac{1}{K}\sum_{r_k \trianglelefteq r_j}v_k^{(l-1)}, l>1 \\
% \end{align}
\begin{equation}
\small
V(r_j) = 
\begin{cases}
% replace \frac with \dfrac{}{} in cases
    v(r_j) + \alpha \circ \bar v (R_{\trianglelefteq r_j}), &  l>1;\\
    % r_k \trianglelefteq r_j
    v(r_j), & l=1.\\
\end{cases}
\end{equation}
where $\circ$ refers to the element-wise product between the centroid of the sub-region descriptive vector cluster $\bar v (r_k)$ and a weighted vector $\alpha$ derived from dimension-wise sparseness. 
\begin{align}
\small
\bar v (R_{\trianglelefteq r_j}) = \frac{1}{|R_{\trianglelefteq r_j}|} \cdot \sum_{r_k \in R_{\trianglelefteq r_j}}v(r_k)
\end{align}
Similar to Eq.~\ref{eq:compact}, we can solely take a dimension in the descriptive vector to calculate the sparseness, which represents the regional bias related to the description word $d_i$.
\begin{align}
\small
c(R_{\trianglelefteq r_j})_i = & \frac{2}{|R_{\trianglelefteq r_j}|(|R_{\trianglelefteq r_j}|-1)} \cdot \\ \notag
& \sum_{r_{k1},r_{k2} \in R_{\trianglelefteq r_j}} ||v(r_{k1})_i-v(r_{k2})_i|| 
\end{align}
As for each specific dimension $i$ in the weighted vector $\alpha$, we use a softmax operation to calculate them:
\begin{align}
\small
     \alpha_{i} = \frac{e^{c(R_{\trianglelefteq r_j})_i}}{\sum_{{i'}=1}^{n} e^{c(R_{\trianglelefteq r_j})_{i'}}}
\end{align}
In short, the aggregated descriptive vector $V$ introduces the information from the lower level by utilising the centroid of the sub-region cluster, while carefully considering the variances among different stereotype descriptions and integrating them with the weighted vector $\alpha$.
% of the current region to represent the general bias information for region $r_j$ :

To introduce the hierarchical information into the measurement of regional bias in language models, we define an aggregation function corresponding to the cluster-based metric described in \S\ref{sec:method_eval_compact_c}, which calculates the bias for region $r_j$ at level $l$.
% \begin{equation}
% \begin{align}
%     % B_{region}(\theta) = & c(R_{city}) + c(R_{country}) + \\   \notag
%     % &  c(R_{continent})    
%     C(r_j) = softmax(V(r_j)) \circ c(R_{\trianglelefteq r_j})
% \end{align}
\begin{equation}\label{eq:C_w}
\small
C_{w}(r_j) = 
\begin{cases}
% replace \frac with \dfrac{}{} in cases
    % \limits_{r_{k1},r_{k2} \in R_{\trianglelefteq r_j}}
     \dfrac{2}{|R_{\trianglelefteq r_j}|(|R_{\trianglelefteq r_j}|-1)} \cdot \sum\limits_{r_{k1},r_{k2} \in R_{\trianglelefteq r_j}} ( &  \\ w_{r_{k1}r_{k2}}\cdot 
     ||V(r_{k1}) -V(r_{k2})||),  & l>1  ; \\
     \\
     ||v(r_j)-\bar v(R_{r_j \trianglelefteq})||, & l=1.
\end{cases}
\end{equation}
where $w_{r_{k1}r_{k2}}$ is a weighted term for the pairwise distance between aggregated descriptive vectors $V$. 
The bias of regions at the lowest level are represented by the distance to their centroids $\bar v$, since there are no sub-regions.
As the aggregated sparseness function should utilise the sparseness of sub-regions, we add the weighted term with respect to the sparseness summation of the sub-regions and formalise it as:
\begin{equation}\label{eq:w}
\small
    w_{r_{k1}r_{k2}} = \frac{e^{C(r_{k1}) + C(r_{k2})}}{\sum\limits_{r_{k1'},r_{k2'} \in R_{\trianglelefteq r_j}}e^{C(r_{k1'}) + C(r_{k2'})}}
\end{equation}
By exploiting the hierarchical architecture of the regional social groups, our evaluation method applies a from-bottom-to-up design to capture the propagation of information. The aggregated sparseness metric provides an intuitive method for the hierarchical regional bias evaluation, with which we can add a root node `the Earth' on the top of the social group hierarchy to represent the whole society and measure the overall bias in language models.

\subsection{Region Probability Weighted Variant}\label{sec:method_eval_region_weight}
As the weighted term in Eq.~\ref{eq:w} is calculated according to the sub-region biases for the aggregated descriptive vectors, we argued that it could be replaced with the contextualised likelihood of the single \texttt{\small [region]} words to leverage the importance learned by the language model in the bias evaluation.
% Using the frequency in corpus to represent the possibility of being biased
We propose to acquire the such a regional likelihood learned by the LMs by passing the single word \texttt{\small [region]} $r_j$ into the Eq.~\ref{eq:f} $f(r_j)$ to approximate the contextualised likelihood of the given region.
\begin{equation}
\small
    z_{r_{k1}r_{k2}} = \frac{e^{f(r_{k1}) + f(r_{k2})}}{\sum\limits_{r_{k1'},r_{k2'} \in R_{\trianglelefteq r_j}}e^{f(r_{k1'}) + f(r_{k2'})}}
\end{equation}
The variant aggregated regional bias measure function is noted as $C_z$, where the $w_{r_{k1}r_{k2}}$ in Eq.~\ref{eq:C_w} is replaced with $z_{r_{k1}r_{k2}}$. In the variant metric $C_z$, hierarchical information is only modelled in the calculations of descriptive vectors.

% \section{Regional Bias Mitigation}

% \section{Experiments}

% % \subsection{Experimental Details}

% \textcolor{blue}{summarise the section}

% \noindent \textbf{Description Words.} 

% \noindent \textbf{Region Words.} 

% \noindent \textbf{Language Models.} 

% \noindent \textbf{Implementation.} 

\begin{table*}[!bt]
% \large
%  \textcolor{red}{verify the results again}
\centering
\scalebox{.85}{
\begin{threeparttable}\small
\begin{tabular}{c|c|cccccc|c} % |cccc
    \hline
    % \multirow{2}{*}{\textbf{Model}} 
   \textbf{Model} & \multirow{2}{*}{\textbf{Metric}} &  \multicolumn{6}{c|}{\textbf{Continent-level Results}} & 
   \textbf{Overall} % \multirow{2}{*}{\textbf{Overall}} 
   \\ 
   % \cline{3-8}
    \text{\scriptsize Parameter, Corpora}  & & AF & AS & EU & OC$^{3^{rd}}$ & SA$^{2^{nd}}$ & NA$^{1^{st}}$ & \textbf{Bias}  \\
    \hline \hline 
    BERT\text{\tiny Base}  & $C_w$ 
    &0.0227
    &0.0283
    &0.0245
    &0.0445
    &0.1061
    &0.3185
    &2.3223\\
    \text{\scriptsize 110M, $\spadesuit$}   & $C_z$ 
    & 0.0227
    & 0.0282
    & 0.0245
    & 0.0444
    & 0.1072
    & 0.3205
    & 2.3271   \\
    % \cline{1-2}\cdashline{3-10}
    \hline

    ALBERT\text{\tiny Base-V2} & $C_w$ 
    &0.0322
    &0.0371
    &0.0372
    &0.0703
    &0.1827
    &0.5152
    &3.3045  \\
    \text{\scriptsize 12M, $\spadesuit$}  & $C_z$ 
    & 0.0322
    & 0.0374
    & 0.0372
    & 0.0701
    & 0.1850
    & 0.5211
    & 3.3150  \\
    % \cline{1-2}\cdashline{3-10}
    \hline
    % DistilBERT\text{\tiny Base}  & $C_w$ 
    % &0.0208
    % &0.0215
    % &0.0199
    % &0.0433
    % &0.0974
    % &0.2343
    % &1.6123  \\
    %  \text{\scriptsize 66M} & $C_z$ 
    %  & 0.0210
    %  & 0.0216
    %  & 0.0195
    %  & 0.0431
    %  & 0.0973
    %  & 0.2407
    %  & 1.5769  \\
    % \hline

    RoBERTa\text{\tiny Base}  & $C_w$ 
    & 0.0437
    & 0.0354
    & 0.0391
    & 0.0848
    & 0.2109
    & 0.5048
    & 3.2274
    \\

     \text{\scriptsize 125M, $\clubsuit$}  & $C_z$  
     & 0.0436
     & 0.0354
     & 0.0391
     & 0.0846
     & 0.2110
     & 0.4984
     & 3.2226  \\
    % \cline{1-2}\cdashline{3-10}
    \hline
    
    BART\text{\tiny Base}& $C_w$ 
    & 0.0073
    & 0.0094 
    & 0.0069 
    & 0.0138 
    & 0.0329 
    & 0.1153 
    & 0.5732  \\
     \text{\scriptsize 140M, $\clubsuit$} & $C_z$ 
    & 0.0072
    & 0.0090 
    & 0.0069 
    & 0.0138 
    & 0.0330 
    & 0.1152 
    & 0.8653  \\
    \hline
\end{tabular}
\begin{tablenotes}
% \scriptsize
\small
  \item * All the statistics are multiplied by $1e3$.
\end{tablenotes}
\end{threeparttable}
}
\caption{Evaluation Results of the Hierarchical Regional Bias (HERB$^{\emojiherb{}{}}$) for Language Models. 
The $\spadesuit$ and $\clubsuit$ mark the same pre-training corpora set used in language model pre-trainings.
The two letter continent abbreviations refer to Africa, Asia, Europe, Oceania, South America, and North America, respectively. NA$^{1^{st}}$, SA$^{2^{nd}}$, and OC$^{3^{rd}}$ suggest that these three continents keep top three biases across all LMs.
}\label{tab:overall_bias}
\end{table*}

\begin{table*}[!bt]
% \large
%  \textcolor{red}{verify the results again}
\centering
\scalebox{0.85}{
\begin{threeparttable}\small
\begin{tabular}{c|cccccc|c} % |cccc
    \hline
   \multirow{2}{*}{\textbf{Model}} &  \multicolumn{6}{c|}{\textbf{Continent-level Results}} & \textbf{Overall} \\ 
%   \cline{3-8}
     & AF & AS & EU & OC & SA & NA & \textbf{Bias}  \\
    
    \hline \hline 
    BERT\text{\tiny Base} 
    &0.0416
    &0.0427
    &0.0439
    &0.0479
    &0.0448
    &0.0413
    &0.0454 \\
    \hline
    ALBERT\text{\tiny Base-V2} 
    &0.0690
    &0.0723
    &0.0747
    &0.0713
    &0.0743
    &0.0775
    &0.0743 \\
    % \cline{1-2}\cdashline{3-10}
    \hline
    % DistilBERT\text{\tiny Base} 
    % &0.0152
    % &0.0154
    % &0.0152
    % &0.0202
    % &0.0314
    % &0.0311
    % & NA\\
    % \hline
    
    RoBERTa\text{\tiny Base} 
    &0.0987
    &0.1038
    &0.1022
    &0.0804
    &0.0895
    &0.1001
    &0.0995 \\
    \hline

    BART\text{\tiny Base}
    &0.0218
    &0.0166
    &0.0181
    &0.0189
    &0.0347
    &0.0168
    &0.0187 \\
    \hline
\end{tabular}

% \begin{tablenotes}
% % \scriptsize
% \small
%   \item * All the statistics are multiplied by $1e-3$.
% \end{tablenotes}
\caption{Non-hierarchical Regional Bias Evaluation with Cluster Sparseness.}\label{tab:compactness_ranking}
\end{threeparttable}
}
\end{table*}
\section{Experiments}
%first conduct regional bias evaluation on different large-scale pre-trained LMs
%trend?
%4.2 validate the effectiveness of cluster-based regional bias evaluation
In this section, we conduct regional bias evaluation on pre-trained language models with the proposed metric HERB$^{\emojiherb{}{}}$.
To validate the design of HERB$^{\emojiherb{}{}}$, we provide a comparison between the aggregated evaluation function and the bias acquired only by cluster sparseness and give an ablation study on the description topics. 
At last, we verify the effectiveness of HERB$^{\emojiherb{}{}}$ by exploring the regional bias before and after the LMs are fine-tuned for the downstream task.
% The comparison experiments
% \textcolor{blue}{summarise the section}

\begin{figure*}[!bt]
\centering
\scalebox{0.85}{
\centering
\begin{subfigure}{0.24\linewidth}
\centering
\includegraphics[width=\linewidth]{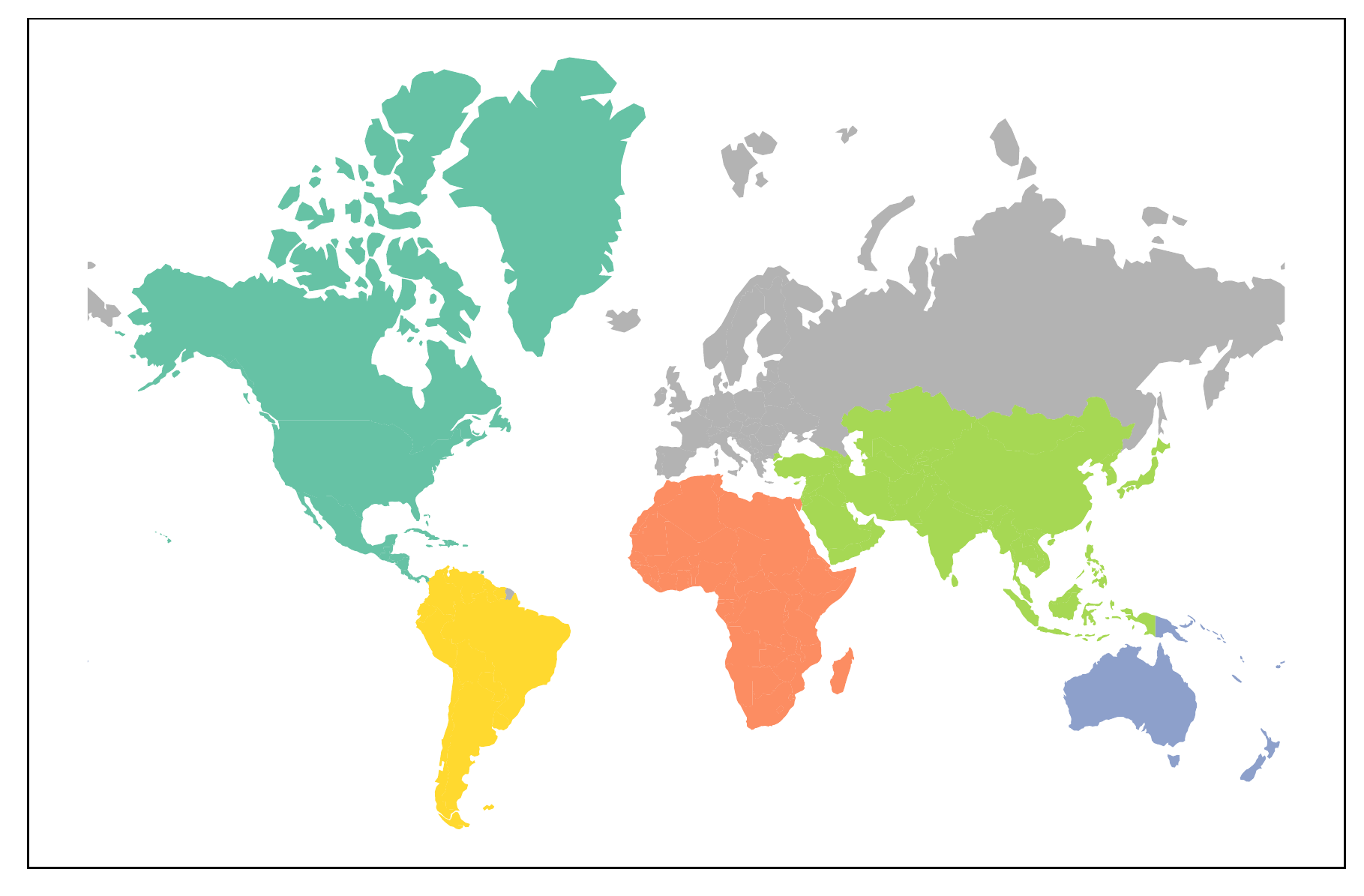} 
\caption{Real  }\label{fig:insight_hier_a}
\end{subfigure}
\begin{subfigure}{0.2\linewidth}
\centering
\includegraphics[width=\linewidth]{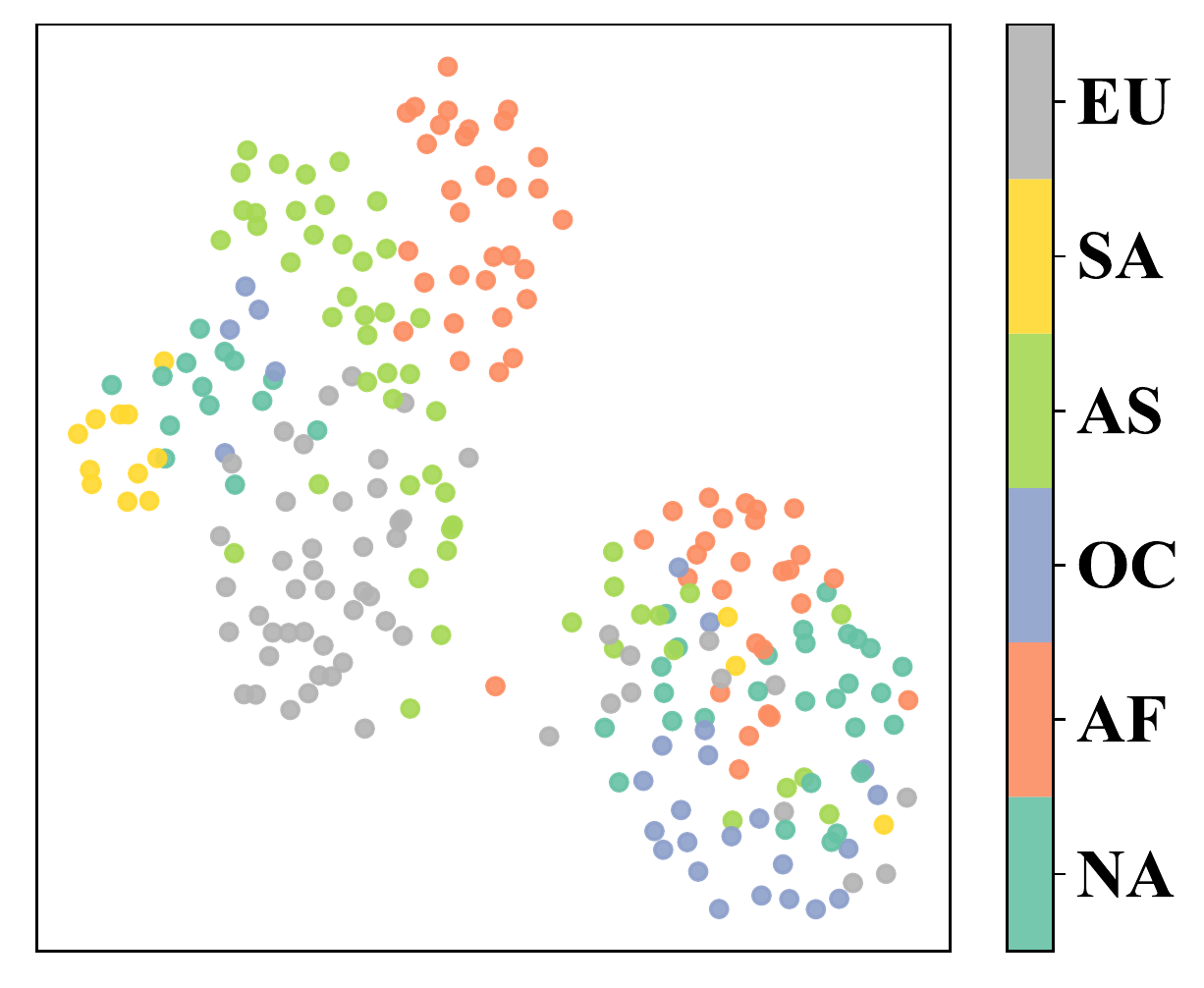} 
\caption{BERT~~~~~~~~~}\label{fig:insight_hier_b}
\end{subfigure}
% \\
\begin{subfigure}{0.2\linewidth}
\centering
\includegraphics[width=\linewidth]{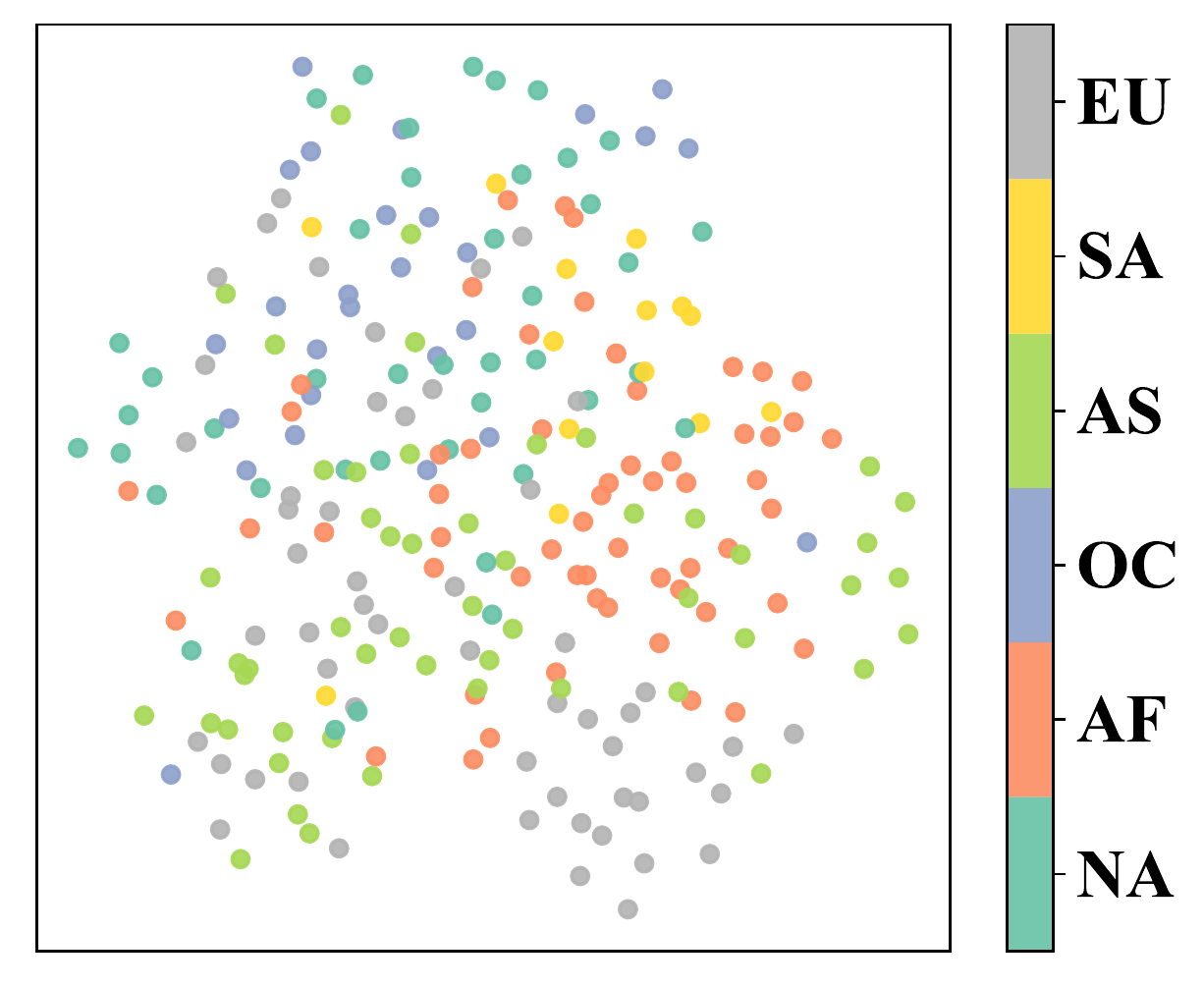} 
\caption{RoBERTa~~~~~~}\label{fig:insight_hier_c}
\end{subfigure}
\begin{subfigure}{0.2\linewidth}
\centering
\includegraphics[width=\linewidth]{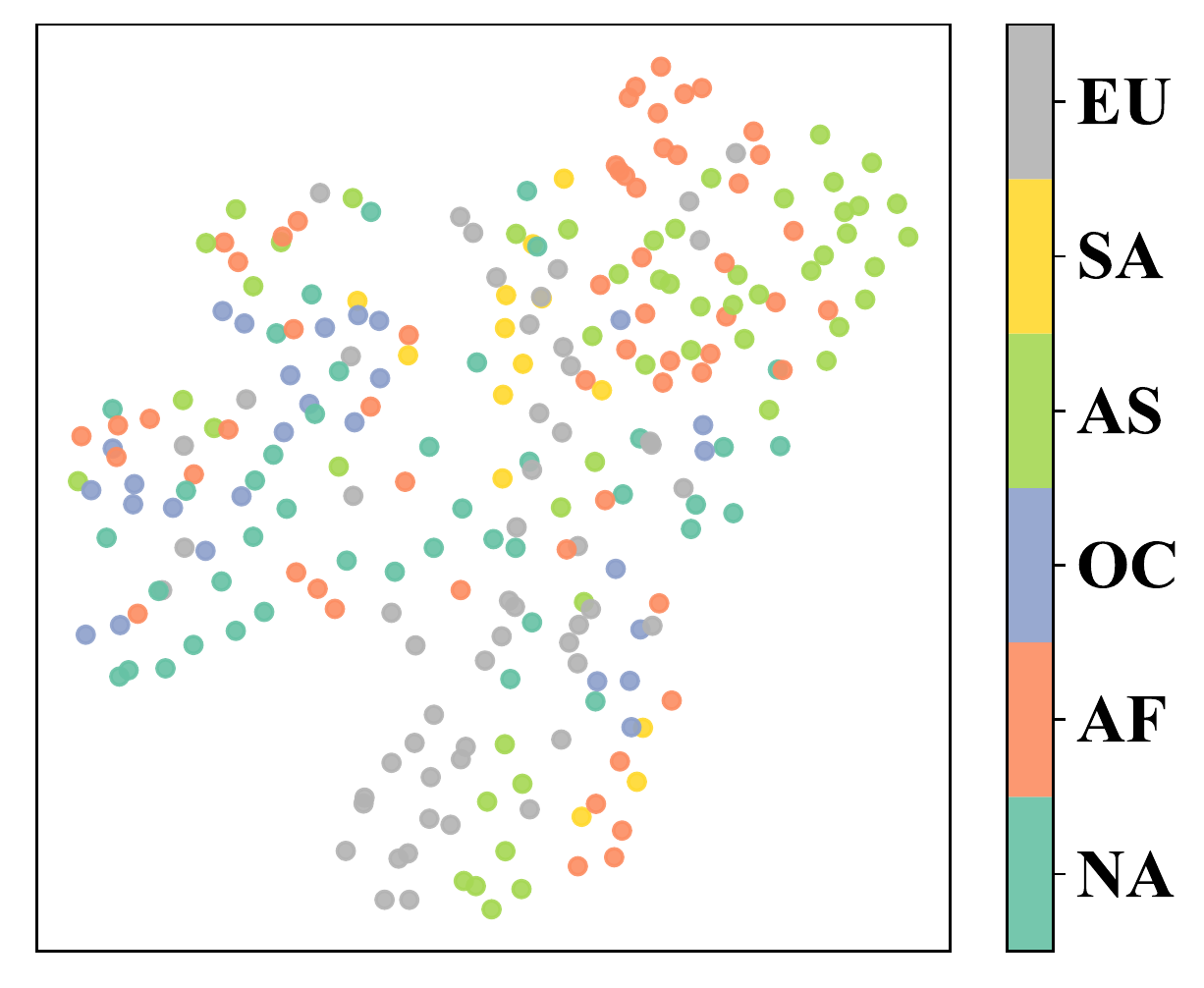} 
\caption{ALBERT~~~~~~}\label{fig:insight_hier_d}
\end{subfigure}
}
\caption{
Distributions of Country-level Regions in the Real World and in the Learned Representation Space. 
% Regions in the Antarctic and uninhabited regions are excluded.
Regions in the Antarctic are excluded.
The plots other than Fig.~\ref{fig:insight_hier_a} are contextualised country representations taken from the learned space of pre-trained language models with the method described in \S\ref{sec:method_eval_region_weight}.
}
\label{fig:insight_hier}
\vspace{-3mm}
\end{figure*}

\subsection{Regional Bias in Pre-trained Models}\label{sec:exp_overall_bias}
\label{sec:LM_Bias}

% 1. 总结现象：
% 1.1 哪个 model 的 overall 最高; 
% 1.2 前三名大洲基本一致
% 2. 解释原因 

%  using the metrics designed in \S\ref{sec:method_eval} 
% We conduct regional bias evaluation on large-scale pre-trained language models including BERT, ALBERT, RoBERTa, and BART~\cite{devlin2018bert, lan2019albert, sanh2019distilbert, liu2019roberta,lewis2019bart}.
% The BERT, ALBERT, and DistilBERT are pre-trained with the same corpora containing the BookCorpus~\cite{moviebook} and English Wikipedia~\cite{wiki:English_Wikipedia}, whilst RoBERTa and BART use BookCorpus, CC-News~\cite{Nagel2016ccnews}, OpenWebText~\cite{gokaslan2019openweb}, and Stories~\cite{trinh2018simple}.
% As shown in Tab.~\ref{tab:overall_bias}, 
We conduct regional bias evaluation on large-scale pre-trained LMs including BERT, ALBERT, RoBERTa, and BART~\cite{devlin2018bert, lan2019albert, liu2019roberta,lewis2019bart} and provide the metrics on the overall bias and biases in continent-levels as shown in Tab.~\ref{tab:overall_bias}.

In the experiments, we discover that ALBERT contains the highest overall regional bias among the selected LMs, followed by RoBERTa, BERT, and BART. 
We hypothesise that the main reason for the low regional bias of BART is that it formulates sentence level
%acquires the ability of sentence 
reasoning in the pre-training. 
Compared to the other LMs, the sentence rotation and document rotation of BART helps the model learn the relationships among sentences rather than only modelling the context witin sentences and distorting it as regional bias.

We also find that the regional bias on different pre-trained LMs holds the same rankings in the two variants of our evaluation methods. Since the variant metrics $C_w$ and $C_z$ differ on the weight of pairwise distance between the aggregated descriptive vectors, the similar results of the variants show that the unchanged aggregated hierarchical descriptive vector $V$ has more impact on the regional bias than the weight strategies.

% After scrutiny of the pre-training settings, we find that the pre-training corpora selections and the model parameter sizes are not the main reasons causing different regional bias scores.
% % of LMs are excluded from the causes of different bias scores since 
% The BERT, ALBERT, and DistilBERT are pre-trained with the same corpora~\cite{moviebook,wiki:English_Wikipedia}, whilst RoBERTa and BART share the same setting~\cite{Nagel2016ccnews, gokaslan2019openweb, trinh2018simple}.
After a scrutiny of the pre-training settings, we find that both the pre-training corpora selections and the model parameter sizes are not the main factors affecting the regional bias scores. 
It can be observed that the language models with similar parameter sizes do not necessarily contain the same level of regional bias, which becomes apparent when comparing the distinguished regional biases of RoBERTa and BART.
Besides, as revealed in Tab.~\ref{tab:overall_bias}, RoBERTa and BART are pre-trained with the same corpora~\cite{moviebook, Nagel2016ccnews, gokaslan2019openweb, trinh2018simple}, whilst BERT and ALBERT apply another setting~\cite{moviebook,wiki:English_Wikipedia}.
This implies that using the same pre-training corpus settings does not guarantee identical regional bias would be integrated into the models.

\subsection{Hierarchy for Cluster-based Bias}

To demonstrate the effectiveness of the designed aggregation functions for the descriptive vectors and cluster-based regional bias, we compare the proposed aggregated regional bias calculation with the plain version defined in Eq.~\ref{eq:v} and Eq.~\ref{eq:compact}, which ignores the hierarchy of regional groups.

We conduct the comparison experiments for the same pre-trained LMs mentioned in \S\ref{sec:exp_overall_bias}.
The plain regional bias evaluation regards all the regions at the same level and acquires the descriptive vector without information from other regional groups.
During the calculation, the plain regional bias puts all the target regional groups into one cluster and models the cluster sparseness by the pairwise $L_2$ distances between the plain descriptive vectors.

As the results revealed in Tab.~\ref{tab:compactness_ranking}, the overall regional bias shows similar tendency with Tab.~\ref{tab:overall_bias}. RoBERTa achieves the highest bias score, followed by ALBERT, BERT and BART.

It is noticeable that the plain regional bias evaluation is not able to enable different LMs to hold the same bias ranking for different continents, e.g. Tab.~\ref{tab:overall_bias} shows LMs allocate North America the highest regional bias score. That is caused by removing the hierarchical group-group information and is crucial for evaluating the overall regional bias. 

% \textcolor{blue}{explanation}

\begin{table*}[!hbt]
% \large

%  \textcolor{red}{verify the results again}
\centering
\scalebox{0.8}{
\begin{tabular}{c|cccccc|c}
    \hline
   \multirow{2}{*}{\textbf{Description}} &  \multicolumn{6}{c|}{\textbf{Continent-level Results}} & \textbf{Overall} \\ 
   % \cline{3-9}
    & AF & AS & EU & OC & SA & NA & \textbf{Bias} \\
    \hline \hline 
        % \row{}{*}{} 
    Full List
    &0.0322
    &0.0371
    &0.0372
    &0.0703
    &0.1827
    &0.5152
    &3.3045  \\
    \hline 
    % \row{}{*}{}
    w/o Occupation
    & 0.0316
    & 0.0372
    & 0.0374
    & 0.0689
    & 0.1801
    & 0.5070
    & 3.3410  \\
    % \cline{1-2}\cdashline{3-10}
    \hline
    % \row{}{*}{} 
    w/o Intelligence
    & 0.0318
    & 0.0365
    & 0.0365
    & 0.0702
    & 0.1800
    & 0.5154
    & 3.2947  \\
    % \cline{1-2}\cdashline{3-10}
    \hline
    % \row{}{*}{}
    w/o Appearance
    & 0.0323
    & 0.0373
    & 0.0383
    & 0.0699
    & 0.1838
    & 0.5201
    & 3.3870\\
    % \cline{1-2}\cdashline{3-10}
    \hline
    % \row{}{*}{} 
    w/o Strength
    & 0.0314
    & 0.0349
    & 0.0353
    & 0.0685
    & 0.1831
    & 0.5035
    & 2.9390
 \\
    % \cline{1-2}\cdashline{3-10}
    \hline
    % \row{}{*}{} 
    w/o Morality
    & 0.0325
    & 0.0378
    & 0.0374
    & 0.0709
    & 0.1807
    & 0.5123
    & 3.3970 \\
    % \cline{1-2}\cdashline{3-10}
    \hline
\end{tabular}
}
\vspace{-2mm}
\caption{Ablation Study of Descriptive Topics with ALBERT.}\label{tab:ablation}
\end{table*}
\begin{figure}[!bt]
\centering
{
\centering
\begin{subfigure}{0.44\linewidth}
\centering
\includegraphics[width=\linewidth]{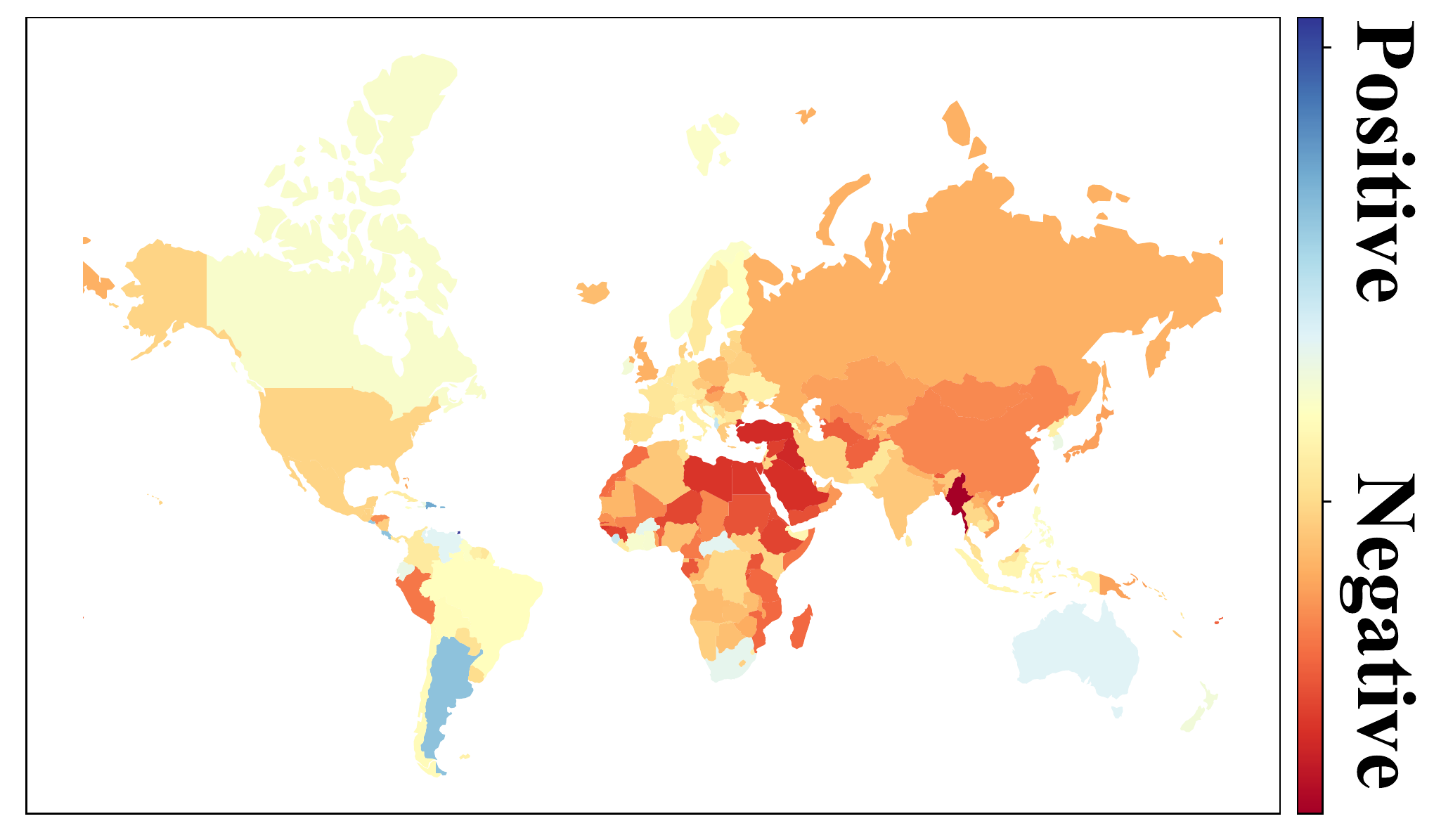} 
\caption{Review Sentiment~~~}\label{fig:task_pred_change_a}
\end{subfigure}
\begin{subfigure}{0.44\linewidth}
\centering
\includegraphics[width=\linewidth]{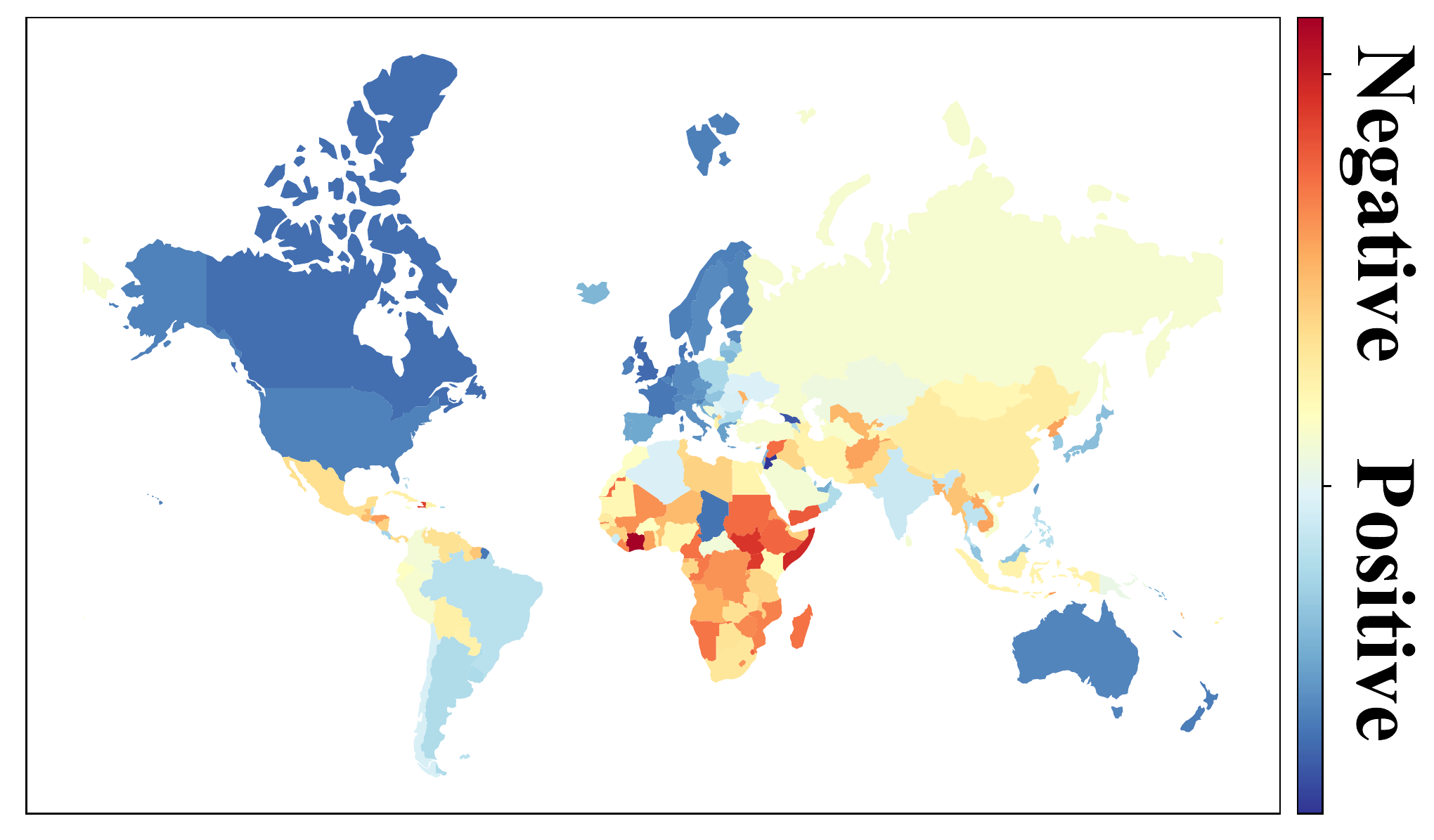} 
\caption{Hate Speech Score~~~~}\label{fig:task_pred_change_b}
\end{subfigure}
}
\caption{
Prediction Difference w.r.t Country-level Regional Bias. Fig.~\ref{fig:task_pred_change_a} and Fig.~\ref{fig:task_pred_change_b} refer to the prediction changes on the sentiment classification task on IMDB reviews and the hate speech detection task on hatespeech18 dataset, respectively. The plot demonstrates the changes of the proportion of positive predictions in the test samples. More details can be referred to \S\ref{sec:exp_application}.
}
\label{fig:task_pred_change}
\end{figure}
% We find that the regional bias on different pre-trained LMs shows the same trends in the two variants.

\subsection{Hierarchical Region Representations}
From the perspective of representation space, we design an experiment to validate the utility of the proposed hierarchical evaluation design for regional bias. 

To demonstrate the regional social group partitioning in the representation space learned by the language models, we compare the actual regional hierarchy and the contextualised representations of the single \texttt{\small [region]} word as described in~\S\ref{sec:method_eval_region_weight}. 
As presented in Fig.~\ref{fig:insight_hier}, we visualise the representations with UMAP and find that countries on the same continent are placed close to each other in the representation space learned by different LMs.
This suggests that the LMs have learned the real-world hierarchical architecture of regional social groups in the pre-training, which again justify the design of our aggregated evaluation functions.

\subsection{Ablation of Descriptive Topics}\label{sec:ablation}

To study the effects of different types of descriptive topics, we conduct an ablation experiment with ALBERT by separately excluding words in the topics of \textit{occupation, intelligence, appearance, strength, and morality}.

Since the descriptive vector $v$ is all normalised, the overall bias would not be directly affected by the reduced dimension number but by the actual bias brought about by the eliminated description words.
As the results demonstrated in Tab.~\ref{tab:ablation}, the overall bias is changed to various extents when the descriptive words are removed.
The removal of words about \textit{strength} and \textit{intelligence} reduces the overall regional bias, which indicates the ALBERT model learns more biases from such two topics. 

\begin{table*}[!hbt]
% \large

%  \textcolor{red}{verify the results again}
\centering
\scalebox{0.75}{
    \begin{tabular}{c|cccc|cccc}
    \hline
    % \multicolumn{1}{c|}{\multirow{2}{*}{\textbf{Metric}}} & 
    \multicolumn{1}{c|}{\multirow{2}{*}{\textbf{Testset}}} & \multicolumn{4}{c|}{\textbf{IMDB}} & \multicolumn{4}{c}{\textbf{hatespeech18}}  \\ \cline{2-9}
    & \multicolumn{4}{c|}{\textbf{Overall Metrics}} & \multicolumn{4}{c}{\textbf{Overall Metrics}} \\
    \hline \hline 
      Original &  \multicolumn{1}{c|}{\multirow{2}{*}{Acc.}} & \multicolumn{1}{c|}{.9280}  & \multicolumn{1}{c|}{\multirow{2}{*}{Marco F1}} & \multicolumn{1}{c|}{.9280}  &  \multicolumn{1}{c|}{\multirow{2}{*}{Acc.}} & \multicolumn{1}{c|}{.8808}  & \multicolumn{1}{c|}{\multirow{2}{*}{Marco F1}}& \multicolumn{1}{c}{.8795} \\
     Country-All & \multicolumn{1}{c|}{} & \multicolumn{1}{c|}{.9270}  & \multicolumn{1}{c|}{} & \multicolumn{1}{c|}{.9270}  &  \multicolumn{1}{c|}{} & \multicolumn{1}{c|}{.8426}  & \multicolumn{1}{c|}{} & \multicolumn{1}{c}{.8396} \\
      \hline \hline
    % \multirow{4}{*}{\textbf{Avg. Score}}
    % \multicolumn{1}{c}{\textbf{Testset}}
    \multirow{2}{*}{\textbf{Testset}} & \multicolumn{4}{c|}{\textbf{Biased Probability Change}} & \multicolumn{4}{c}{\textbf{Biased Probability Change}} \\ \cline{2-9}
     & Quantity$\uparrow$ & Avg. Prob.$\uparrow$ & Quantity$\downarrow$ & Avg. Prob.$\downarrow$ & Quantity$\uparrow$ & Avg. Prob.$\uparrow$ & Quantity$\downarrow$ & Avg. Prob.$\downarrow$ \\
     % \cline{1-9}
     \hline
    Ireland & 13020 & .0177 & 11980 & .0177 & 48 & .0294 & 430 & .0406  \\
    % United Kingdom & - & - & - & - & 48 & .0294 & 430 & .0406  \\
    % Japan & - & - & - & - &  103 & .0160 & 375 & .0460   \\
    Mexico & 11748 & .0166 & 13251 & .0181 & 228 & .0311 & 250 & .0336 \\ 
    Uganda & 10123 & .0156 & 14877 & .0199 &  327 & .0467 & 151 & .0370   \\
    Syria & 9854 & .0155 & 15146 & .0200 &  299 & .0348 & 179 & .0405   \\
    % Haiti & - & - & - & - &  325 & .0405 & 153 & .0332  \\
    % \hdashline
    \hline
 	Irapuato & 10976 & .0174 & 14024 & .0174 &  80  & .0503 & 398 & .0288 \\
    Puebla & 10405 & .0184 & 14595 & .0167 & 93 & .0524  & 385 & .0276 \\
   	Tapachula  &10750 & .0174 & 14250 & .0174 & 139 & .0448  & 339 & .0273 \\
    Mexico-City & 12911 &.0155 & 12089 & .0194 & 160 & .0395 & 318 & .0288 \\
    \hdashline
 	Irapuato, Mexico & 13075 & .0157 & 11925 & .0193 & 247 & .0282 & 231 & .0369 \\
    Puebla, Mexico & 12909 & .0156 & 12091 & .0194 & 117 & .0429 & 361 & .0290 \\
   	Tapachula, Mexico  & 12445 & .0160 & 12554 & .0188 & 259 & .0286 & 219 & .0369 \\
    Mexico-City, Mexico  & 13020 & .0155 & 11979 & .0194 & 140 & .0396 & 338 & .0294 \\
    % City-GB & - & - & - & - & 32.2k & .0341 & 360.2k & .0520   \\
    % City-JP & - & - & - & - & 40.2k & .0167 & 315.9k & .0612    \\
    % City-MX & - & - & - & - & 82.3k & .0199 & 215.9k & .0580  \\
    % \multirow{2}{*}{\textbf{Avg. Score}} & - & - & -  \\
     %  & - & - & -  \\
     \hline
    \end{tabular}
}
\vspace{-2mm}
\caption{
Regional Bias in Existing NLP Applications. 
The prediction results on the test group Country-All refer to all the test samples modified by country-level biases.
% \textcolor{blue}{movie name? find out why Greenland movies are bad.}
}
% The country codes GB, JP, and MX refer to the countries United Kingdom, Japan, and Mexico, correspondingly. The results of the test group Country-All and Cities are merged before calculation.
\label{tab:nlp_app_bias}
\end{table*}
\begin{table*}[!hbt]
% \large

%  \textcolor{red}{verify the results again}
\centering
\scalebox{0.75}{
    \begin{tabular}{c|cccc|cccc|cccc}
    \hline
    % \multicolumn{1}{c|}{\multirow{2}{*}{\textbf{Metric}}} & 
    \multicolumn{1}{c|}{\multirow{2}{*}{\textbf{Testset }}} & \multicolumn{12}{c}{\textbf{Regional Biased Type}} \\ \cline{2-13}
    & \multicolumn{4}{c|}{\textbf{w/o Ireland}} & \multicolumn{4}{c|}{\textbf{w/o Mexico}} &  \multicolumn{4}{c}{\textbf{Country-All Average}}\\
    \hline \hline 
    % \multicolumn{1}{c|}{BERT\text{\small Base}*} 
    % & \multicolumn{1}{c|}{Acc.} & \multicolumn{1}{c|}{.9626}  & \multicolumn{1}{c|}{Marco F1} & \multicolumn{1}{c|}{.9626}  
    % &  \multicolumn{1}{c|}{Acc.} & \multicolumn{1}{c|}{.9626}  & \multicolumn{1}{c|}{Marco F1} & \multicolumn{1}{c|}{.9626} 
    % & \multicolumn{1}{c|}{Acc.} & \multicolumn{1}{c|}{.9626}  & \multicolumn{1}{c|}{Marco F1} & \multicolumn{1}{c}{.9626} \\

    % \multicolumn{1}{c|}{ALBERT\text{\small Base-V2}*} & \multicolumn{1}{c|}{Acc.} & \multicolumn{1}{c|}{.8938}  & \multicolumn{1}{c|}{Marco F1} & \multicolumn{1}{c|}{.8931}  &  
    % \multicolumn{1}{c|}{Acc.} & \multicolumn{1}{c|}{.8937}  & \multicolumn{1}{c|}{Marco F1} & \multicolumn{1}{c|}{.8930} &
    % \multicolumn{1}{c|}{Acc.} & \multicolumn{1}{c|}{.8938}  & \multicolumn{1}{c|}{Marco F1} & \multicolumn{1}{c}{.8931} \\

    % \multicolumn{1}{c|}{DistilBERT\text{\small Base}*} 
    % & \multicolumn{1}{c|}{Acc.} & \multicolumn{1}{c|}{.8822}  & \multicolumn{1}{c|}{Marco F1} & \multicolumn{1}{c|}{.8810}  
    % &  \multicolumn{1}{c|}{Acc.} & \multicolumn{1}{c|}{.8820}  & \multicolumn{1}{c|}{Marco F1} & \multicolumn{1}{c|}{.8808 } 
    % & \multicolumn{1}{c|}{Acc.} & \multicolumn{1}{c|}{.8822}  & \multicolumn{1}{c|}{Marco F1} & \multicolumn{1}{c}{.8809} \\
    % \hline
    
    \multirow{2}{*}{\textbf{Model}} & \multicolumn{4}{c|}{\textbf{Prediction Label Change (\%)}} & \multicolumn{4}{c}{\textbf{Prediction Label Change (\%)}}  & \multicolumn{4}{c}{\textbf{Prediction Label Change (\%)}} \\ \cline{2-13}
     & \multicolumn{2}{c}{\textit{nohate}$\rightarrow$\textit{hate}} &  \multicolumn{2}{c|}{\textit{hate}$\rightarrow$\textit{nohate}}  
     &\multicolumn{2}{c}{\textit{nohate}$\rightarrow$\textit{hate}} &  \multicolumn{2}{c|}{\textit{hate}$\rightarrow$\textit{nohate}}
     &\multicolumn{2}{c}{\textit{nohate}$\rightarrow$\textit{hate}} &  \multicolumn{2}{c}{\textit{hate}$\rightarrow$\textit{nohate}} \\
     \hline
     
     \multicolumn{1}{c|}{BERT\text{\small Base}*}
     & \multicolumn{2}{c}{0.0723} &  \multicolumn{2}{c|}{1.3632}
     & \multicolumn{2}{c}{0.0723} &  \multicolumn{2}{c|}{1.3692}
     & \multicolumn{2}{c}{0.0720} &  \multicolumn{2}{c}{1.3645} \\
     
     \multicolumn{1}{c|}{ALBERT\text{\small Base-V2}*}
     & \multicolumn{2}{c}{1.7944} &  \multicolumn{2}{c|}{4.7301}
     & \multicolumn{2}{c}{1.7901} &  \multicolumn{2}{c|}{4.7360}
     & \multicolumn{2}{c}{1.7914} &  \multicolumn{2}{c}{4.7296} \\

     % \multicolumn{1}{c|}{DistilBERT\text{\small Base}*}
     % & \multicolumn{2}{c}{0.3878} &  \multicolumn{2}{c|}{5.9870}
     % & \multicolumn{2}{c}{0.3878} &  \multicolumn{2}{c|}{6.0185}
     % & \multicolumn{2}{c}{0.3871} &  \multicolumn{2}{c}{6.0000} \\
     
     \multicolumn{1}{c|}{RoBERTa\text{\small Base}*}
     & \multicolumn{2}{c}{0.3325} &  \multicolumn{2}{c|}{4.9376}
     & \multicolumn{2}{c}{0.3300} &  \multicolumn{2}{c|}{4.9452}
     & \multicolumn{2}{c}{0.3312} &  \multicolumn{2}{c}{4.9396} \\
     
     \multicolumn{1}{c|}{BART\text{\small Base}*}
     & \multicolumn{2}{c}{1.0137 } &  \multicolumn{2}{c|}{1.2943}
     & \multicolumn{2}{c}{1.0129} &  \multicolumn{2}{c|}{1.2978}
     & \multicolumn{2}{c}{1.0121} &  \multicolumn{2}{c}{1.2959} \\
     \hline
    
    \end{tabular}
}
\vspace{-2mm}
\caption{
Prediction Change Brought by Regional Bias in Downstream Task. All the performances are from the language models fine-tuned on the hatesppeech18 dataset. The country-all column contains the average changed ratio of predicted labels across all the countries. The `w/o' represents that the modification w.r.t to the specific country is not included in the testset.
}
\label{tab:nlp_bias_dif}
\end{table*}
% \begin{threeparttable}
% \begin{tablenotes}
% % \scriptsize
% \small
%   \item * All the statistics are multiplied by $1e3$.
% \end{tablenotes}
% \caption{Evaluation Results of the Hierarchical Regional Bias for Language Models. 
% The $\spadesuit$ and $\clubsuit$ mark the same pre-training corpora set used in language model pre-trainings.
% }\label{tab:overall_bias}
% \end{threeparttable}

\subsection{Regional Bias in NLP Applications}\label{sec:exp_application}
% 1. 实验描述
% 2. 验证 propagation 到下游
% 3. propagation 后的 regional bias 趋势和 fine-tuned 之前一样，说明我们的 metric 可以帮助预测传递到下游的 regional bias
% 4. 可以传递的原因可能是 propagation 后的 bias 也呈现 hierarchical，说明 metric 设计合理

To verify the propagation of the regional bias in the language models, we propose an experiment to introduce extra region information into the test samples in those tasks where the LMs are skilled in.
% To explore the regional biases in the existing NLP applications, 
We select the binary sentiment classification task on the IMDB movie review dataset~\cite{maas2011IMDBdataset} as well as the hate speech detection task proposed in the hatespeech18 dataset~\cite{gibert2018hate}.
We first conduct regional bias analysis on the public available state-of-the-art language models\footnote{Fine-tuned models are publicly available for the \href{https://huggingface.co/lvwerra/distilbert-imdb}{review-sentiment} and the \href{https://huggingface.co/Narrativaai/deberta-v3-small-finetuned-hate_speech18}{hate-speech} tasks.}.
We design simple prompts as prefixes to add the regional noise information to the test samples in the two datasets:
\vspace{-1mm}
\begin{itemize}[noitemsep]
\small
    \item IMDB: The cast is from \texttt{[region]}. 
    \item hatespeech18: I am from \texttt{[region]}. 
\end{itemize}
\vspace{-1mm}
% as revealed in Tab.~\ref{tab:task_prefix}. 
% \input{Tab/task_prefix.tex}
% In our experiments exploring regional bias in NLP applications, 
The regional bias fine-tuned LMs  contain can thus be represented by the ratio of prediction results that are changed. 
We give the results and change ratio on the country-level biased test set in Tab.~\ref{tab:nlp_app_bias} and plot corresponding prediction probability difference on a map in Fig.~\ref{fig:task_pred_change}.

When regional identities are given, the language models have worse performances on both tasks and intend to produce biases, i.e. changing the original predicted results on different countries in different ways.
For instance, the hate speech detection model generally increases the probability of hate speech prediction when adding `I am from Mexico' as a prefix than `I am from USA', where only the country name varies. 
This implies that the fine-tuned LMs produce different results even though the regional information should be neutral.
% This suggests that even though the regional affiliation of the speaker should be a neutral context for the prediction tasks, the fine-tuned language model for a specific task still produces inconsistent judgements on different regions. 

We then fine-tune the pre-trained LMs measured by our metrics and provide their performances on the noise test set in Tab~\ref{tab:nlp_bias_dif}. 
The overall change of the prediction results shows that the language models have similar bias rankings in the downstream task as retrieved in~\S\ref{sec:exp_overall_bias}, which shows that our evaluation metric can be a reference for the potential regional bias in the fine-tuned language models for downstream tasks.
We argue that the difference between the rankings before and after fine-tuning could be caused by the instability in the LMs.
%, i.e. biases on different regional social groups, as shown  
% We argue that the LMs inherit such regional bias mostly from the pre-trained corpus since the fine-tuned task.
% Para3: regional bias is hierarchical. 
% 3.1 neighbourhood effects 3.2 propagation to upper-level

As revealed in Fig.~\ref{fig:task_pred_change_b}, the language model assigns higher hate speech probabilities to given sentences when it is informed that the speakers are from African countries compared to European ones.
The revealed country-level regional biases share a generally similar trend in the close regions that can be grouped by geographical features, which rationalises the hierarchical design of our metric from the perspective of the downstream task.
We argue that this is because the common linguistic, cultural, and other objective characteristics shared by people in neighbouring regions are distorted into biases during the language model pre-training.
This suggests that the regions in the same cluster can thus be further modelled by our aggregated function, which summarises the bias in higher-level groups.
% , which suggests that the inherent hierarchical structure of the region groups could be significant to the evaluation of regional bias.
% \textcolor{blue}{TODO: analysis on the topics.} % TODO 为什么离得近就会有相似的bias呢？是否可以通过 d-vector 的不同topic维度分析？

% We argue that the biases on other regions belonging to the same upper-level region should be carefully taken into consideration.
% if an additional fine-grained city names are given, the models will produce .

% \subsection{Regional Bias Mitigation}
% \input{Tab/debias_result.tex}
% Tab.~\ref{tab:debias_result} 
\begin{table*}[!hbt]
% \large
\
%  \textcolor{red}{verify the results again}
\centering
\scalebox{0.75}{
\begin{tabular}{c|cccccc|c}
    \hline
   \multirow{2}{*}{\textbf{Description}} &  \multicolumn{6}{c|}{\textbf{Continent-level Results}} & \textbf{Overall} \\ 
   % \cline{3-9}
    & AF & AS & EU & OC & SA & NA & \textbf{Bias} \\
    \hline \hline 
        % \row{}{*}{} 
    Full List
    &0.0322
    &0.0371
    &0.0372
    &0.0703
    &0.1827
    &0.5152
    &3.3045  \\
    \hline 
    % \row{}{*}{}
    Replace Occupation
    & 0.0330
    & 0.0382
    & 0.0388
    & 0.0721
    & 0.1857
    & 0.5315
    & 3.4786  \\
    % \hline
    % w/o Occupation
    % & 0.0316
    % & 0.0372
    % & 0.0374
    % & 0.0689
    % & 0.1801
    % & 0.5070
    % & 3.3410  \\
    \hline
    % \row{}{*}{} 
    Replace Intelligence
    & 0.0335
    & 0.0376
    & 0.0373
    & 0.0716
    & 0.1835
    & 0.5438
    & 3.2152  \\
    % \hline
    % w/o Intelligence
    % & 0.0318
    % & 0.0365
    % & 0.0365
    % & 0.0702
    % & 0.1800
    % & 0.5154
    % & 3.2947  \\
    \hline
    % \row{}{*}{}
    Replace Appearance
    & 0.0349
    & 0.0400
    & 0.0403
    & 0.0740
    & 0.1953
    & 0.5688
    & 3.3734\\
    % \hline
    % w/o Appearance
    % & 0.0323
    % & 0.0373
    % & 0.0383
    % & 0.0699
    % & 0.1838
    % & 0.5201
    % & 3.3870\\    
    \hline
    % \row{}{*}{} 
    Replace Strength
    & 0.0341
    & 0.0380
    & 0.0379
    & 0.0739
    & 0.1907
    & 0.5323
    & 3.2607
 \\
    %  \hline
    % \row{}{*}{} 
    % w/o Strength
    % & 0.0314
    % & 0.0349
    % & 0.0353
    % & 0.0685
    % & 0.1831
    % & 0.5035
    % & 2.9390\\
    % \cline{1-2}\cdashline{3-10}
    \hline
    % \row{}{*}{} 
    Replace Morality
    & 0.0341
    & 0.0396
    & 0.0389
    & 0.0737
    & 0.1900
    & 0.5403
    & 3.4558
 \\
    %  \hline

    %  w/o Morality
    % & 0.0325
    % & 0.0378
    % & 0.0374
    % & 0.0709
    % & 0.1807
    % & 0.5123
    % & 3.3970 \\
    \hline
    % \cline{1-2}\cdashline{3-10}

\end{tabular}
}
\vspace{-2mm}
\caption{Robustness Study of Descriptive Topic Words with ALBERT.}\label{tab:ablationNew}
\end{table*}
\begin{figure*}[!bt]
\centering
\includegraphics[width=0.75\linewidth]{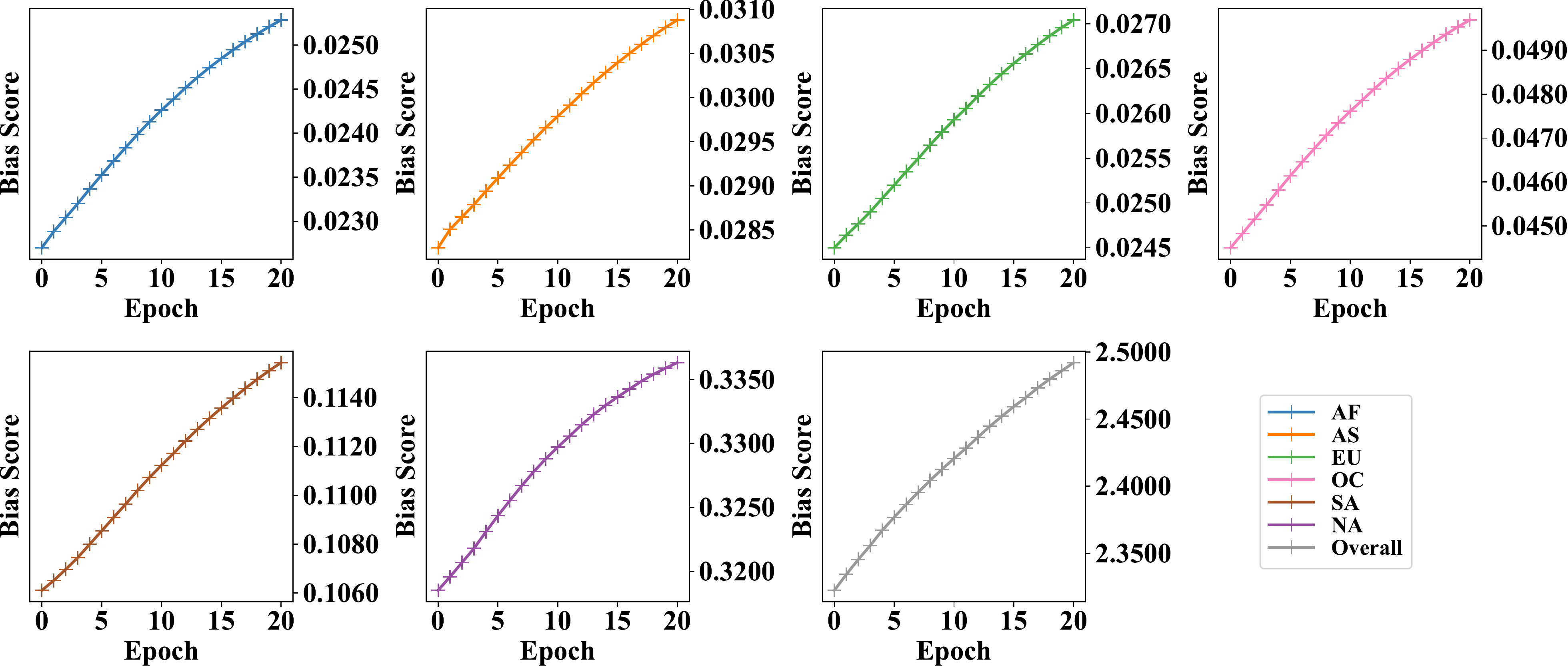} 
\caption{
HERB$^{\emojiherb{}{}}$ Evaluation on BERT along the Toxic MLM Training Task. The overall regional bias and continent-level bias scores (multiplied by 1e3) of the model are plotted separately.
}\label{fig:toxic_MLM}
\vspace{-2mm}
\end{figure*}

\subsection{Robustness Study for Word Choice}\label{sec:result_ablation}
% In addition to evaluate our metric's robustness by removing description word in each topic individually, we further evaluate the sensitivity of word choices in each topic as \citet{antoniak-mimno-2021-bad} indicates that bias metrics can potentially be unreliable to changes in word choices.
\citet{antoniak-mimno-2021-bad} suggests that bias metrics may be potentially unreliable to changes in word choices, thus we further analyze the sensitivity of word choices in each topic in addition to evaluating the robustness of our metric by eliminating description words from each topic separately.
We design an experiment to evaluate the HERB$^{\emojiherb{}{}}$ of ALBERT while replacing the descriptive words in one of the topic.

We first calculate the most similar word for each descriptive word in Appendix~\ref{sec:appendix_wordlist} with the word embedding method. 
Then we conduct a robustness testing experiment with ALBERT by separately replacing words in the topics of \textit{occupation, intelligence, appearance, strength, and morality}. Then the regional bias calculated with the accordingly derived five description word list are calculated. 

As the results demonstrated in Tab.~\ref{tab:ablationNew}, we notice that resultant biases do not differ much from the initial overall bias when the descriptive words are replaced. Even though word choices fluctuate, our evaluation metric' results stay consistent, proving the robustness and reliability of HERB$^{\emojiherb{}{}}$.

\subsection{Interpreting the HERB$^{\emojiherb{}{}}$ Score}\label{sec:appendix_score_interpreting}

Although the HERB$^{\emojiherb{}{}}$ scores already provide a guidance to audit and compare the regional bias among different PLMs, we conduct an additional experiment to further quantify the scores and improve the intuitive interpretation of the evaluation report. We design a toxic corpus masked language modelling (MLM) task for continual training on the pre-trained BERT, which feeds toxic regional-biased sentences into the model.
% and trains the model to predict the masked region words.

We construct the toxic corpus with template sentences that get top-20 values calculated by Eq.~\ref{eq:f} regards to each description word, which results in total 2240 sentences. 
We then mask the regional information of the sentences and train the model with MLM task.
To illustrate the affect from the toxic corpus best, the model is trained with simple SGD optimiser~\cite{Robbins2007ASA_SGD} and constant learning rate $5e-5$ for 20 epochs.

The model is saved and evaluated after each epoch during the toxic MLM training. As shown in Fig.~\ref{fig:toxic_MLM}, the overall and continent-level biases show positive correlation to the number of train epochs. 
Since the bias score increases as more toxic sentences are fed, HERB$^{\emojiherb{}{}}$ shows the ability to reflect the quantity of biased corpus integrated into the LMs during the pre-training.
\section{Conclusion}
% \textcolor{blue}{
% 1. 没有人做过、
% 2. Hierarchical 符合regional 的特点
% 3. 丰富的词表
% 4. 为下游任务选择 pre-trained model 的时候， 从 fair 的角度提供建议
% }

In this work, regional bias in the pre-trained language models has been measured in depth for the first time within the NLP community. 
% first time leaps into the consciousness of the NLP community.
% To our best knowledge, we are the first to analyse the regional bias in LMs, and additionally propose metrics specifically for it.
The proposed metric, HERB$^{\emojiherb{}{}}$, takes hierarchical characteristics of regional bias into consideration and adopts a carefully selected descriptive word list.
% coming from previous research works after slight modification.
We use HERB$^{\emojiherb{}{}}$ to evaluate regional bias in state-of-the-art language models and validate the robustness of HERB$^{\emojiherb{}{}}$ by providing bias analysis on downstream tasks for corresponding models. 
% which sheds light on the selection of pre-trained model with respect to regional bias for downstream tasks.
Thorough experimentation studies are given to show that the hierarchical structure of regions does not only present in the pre-trained representation space but also appears as hierarchical bias in downstream tasks, which further rationalises the design of HERB$^{\emojiherb{}{}}$.
% regional bias propagating downstream and verify the existence of 
% , investigate the influence of different components of our metrics, and provide references for downstream backbone LMs' selection from the regional bias aspect.
% In conclusion, we propose an evaluation method to quantify the regional bias in pre-trained LMs, while considering the hierarchical regional group structure and measuring the bias with selective description words.
% from a wide range of aspects defined by selective description words.

\clearpage
\section*{Limitations}
% 只有metrics，没有解决方案
% 没有对corpus来源造成的bias score的变化的系统分析
% 证明了我们的score和下游任务以及包含的bias之间的正相关关系，但是没有对随着包含的bias的增长我们的score会怎么变化的量化分析。
In our work, we only provide a hierarchical evaluation for LMs targeting the regional social groups but not corresponding mitigation methods for such contextualised bias. 
We argue that the hierarchical structure should also be considered in the regional bias mitigation due to the natural geographical clusters learned in the models, which could be placed into another work for the completeness of presentation.
Moreover, although we try to provide short and simple template for contextual encoding in the evaluation, the template may not cover all the aspects of the identification of the speaker. This could be further explored by localising the expressions for different regional identification, which may benefit the effectiveness of bias evaluation.
% Another analysis also limited is the quantified study between the final score
% One other possible further quantified analysis for the 
% In our experiments, we found that the continent-wise regional bias measured by our metric shows a negative correlation of the number of countries/regions contained inside the continent. 
% Even though we have put efforts into eliminating such an influence by dividing the number of regions each time when the descriptive vectors and regional cluster sparseness are aggregated, we can not completely remove it.
% One possible refinement of our design is hence to normalise the number of regional groups in the evaluation.
\section*{Acknowledgement}
Yizhi Li is fully funded by an industrial PhD studentship (Grant number: 171362) from the University of Sheffield, UK. This work is also  supported by the National Key R\&D Program of China (2020AAA0105200).
% \newpage
% \clearpage

\bibliographystyle{acl_natbib}
\bibliography{reference.bib}

\clearpage
\appendix

\section{Description Word List}\label{sec:appendix_wordlist}

We provide the detailed full description word list in the following five categories.

\noindent \textbf{Occupation} homemaker, nurse, receptionist, librarian, socialite, hairdresser, nanny, bookkeeper, stylist, housekeeper, maestro, skipper, protege, philosopher, captain, architect, financier, warrior, broadcaster, magician

\noindent \textbf{Intelligence} precocious, resourceful, inquisitive, genius, inventive, astute, adaptable, reflective, discerning, intuitive, inquiring, judicious, analytical, apt, venerable, imaginative, shrewd, thoughtful, wise, smart, ingenious, clever, brilliant, logical, intelligent

\noindent \textbf{Appearance} alluring, voluptuous, blushing, homely, plump, sensual, gorgeous, slim, bald, athletic, fashionable, stout, ugly, muscular, slender, feeble, handsome, healthy, attractive, fat, weak, thin, pretty, beautiful, strong

\noindent \textbf{Strength} powerful, strong, confident, dominant, potent, command, assert, loud, bold, succeed, triumph, leader, dynamic, winner, weak, surrender, timid, vulnerable, wispy, failure, shy, fragile, loser

\noindent \textbf{Morality} upright, honest, loyal, gentle, treacherous, clownish, brave, kind, hard-working, thrifty, optimistic, tolerant, earnest, straightforward, narrow-minded, humble, punctual, single-minded, uncompromising

\section{Substituted Description Word List}\label{sec:appendix_wordlist_sub}
We provide the detailed full substitution description word list in the following five categories, each word in most similar word calculated by word embedding method.
\noindent \textbf{Occupation} 
housewife,
doctor,
waitress,
archivist,
businesswoman,
manicurist,
housekeeper,
janitor,
stylists,
nanny,
virtuoso,
captain,
protégé,
mathematician,
skipper,
sculptor,
billionaire,
dragon,
television,
illusionist

\noindent \textbf{Intelligence} 
gawky,
industrious,
perceptive,
visionary,
imaginative,
shrewd,
resourceful,
textured,
jaded,
instinctive,
enquiring,
diligent,
methodology,
ironic,
storied,
inventive,
canny,
insightful,
good,
intelligent,
inventive,
clumsy,
superb,
rational,
smart

\noindent \textbf{Appearance} seductive,
curvaceous,
wrinkling,
geeky,
scrawny,
sensuous,
lovely,
slimmer,
eagle,
basketball,
trendy,
slender,
nasty,
skeletal,
elongated,
anemic,
charming,
healthier,
desirable,
calories,
weaker,
thick,
quite,
lovely,
stronger

\noindent \textbf{Strength} strong,
stronger,
optimistic,
predominant,
powerful,
commander,
asserting,
deafening,
daring,
successor,
victory,
party,
interaction,
winners,
weaker,
surrendered,
hesitant,
susceptible,
spiky,
failed,
timid,
shaky,
losers

\noindent \textbf{Morality} sturdy,
truthful,
loyalists,
playful,
perilous,
buffoonish,
courageous,
sort,
hardworking,
frugal,
pessimistic,
intolerant,
thoughtful,
simple,
self-important,
unassuming,
courteous,
monomaniacal,
unyielding

\end{document}